\documentclass[10pt,twocolumn,letterpaper]{article}

\usepackage{cvpr}
\usepackage{times}
\usepackage{epsfig}
\usepackage{graphicx}
\usepackage{amsmath}
\usepackage{amssymb}

\usepackage{multirow}

\usepackage{bm}
\usepackage{array}
\usepackage{color}

\usepackage{movie15}

\usepackage[pagebackref=true,breaklinks=true,letterpaper=true,colorlinks,bookmarks=false]{hyperref}

\cvprfinalcopy 

\def\cvprPaperID{2915} 

\begin{document}

\title{Face Translation between Images and Videos using Identity-aware CycleGAN}

\author{Zhiwu Huang$^\dagger$, Bernhard Kratzwald$^\dagger$, Danda Pani Paudel$^\dagger$, Jiqing Wu$^\dagger$, Luc Van Gool$^{\dagger\ddagger}$\\
	$^\dagger$Computer Vision Lab, ETH Zurich, Switzerland \quad $^\ddagger$VISICS, KU Leuven, Belgium\\
	{\tt\small \{zhiwu.huang, paudel, jwu, vangool\}@vision.ee.ethz.ch, bkratzwald@ethz.ch}}
\maketitle

\begin{abstract}
	This paper presents a new problem of unpaired face translation between images and videos, which can be applied to facial video prediction and enhancement. In this problem there exist two major technical challenges: 1) designing a robust translation model between static images and dynamic videos, and 2) preserving facial identity during image-video translation. To address such two problems, we generalize the state-of-the-art image-to-image translation network (Cycle-Consistent Adversarial Networks) to the image-to-video/video-to-image translation context by exploiting a image-video translation model and an identity preservation model. In particular, we apply the state-of-the-art Wasserstein GAN technique to the setting of image-video translation for better convergence, and we meanwhile introduce a face verificator to ensure the identity. Experiments on standard image/video face datasets demonstrate the effectiveness of the proposed model in both terms of qualitative and quantitative evaluations. 
\end{abstract}

\section{Introduction}

In recent years, face translation between images has attracted a lot of attention. For example, \cite{huang2017beyond, li2017generative, tran2017disentangled, li2017generate} translate facial images with identity-preserving. However, few works make efforts to translate faces between images and videos. To fill this gap, we propose the problem of bidirectional face translation between images and videos. Actually, image-video face translation has a broad range of applications. On one hand, given a static facial image, predicting its dynamic pattern in videos is very useful. On the other hand, it is also in high demand for video enhancement by translating low quality videos to high quality images. 

Generally, image-video face translation suffers from more challenges than image-based face translation. One of the most challenging issues is treating the high heterogeneity between static images and dynamic videos. This challenge will become more complex when there are merely a limited number of paired image-video faces which belong to the same identity. Similar to the traditional image-to-image translation task, it is also really difficult for us to collect paired images and videos as noted in \cite{zhu2017unpaired}. As image-based face translation, the other major issue is that it is non-trivial to maintain the identity information especially when the involved faces are from two heterogeneous signals, that are 2D images and 3D videos. 

With such motivation in mind, this paper focuses on the challenging problem of unpaired face translation between images and videos. In order to address the two crucial issues aforementioned, we aim to propose a bidirectional image-video face translation model in this paper. Inspired by the state-of-the-art image-to-image translation model (Cycle-Consistent Adversarial Networks, CycleGAN) \cite{zhu2017unpaired}, we design a cyclic image-video translation network with exploiting a image-to-video translation model, a video-to-image translation model, an image critic model, a video critic model and an identity preserving model. The image-to-video translation model consists of a pair of image encoder and video decoder, while the video-to-image translation model contains a pair of video encoder and image decoder. In addition to exploiting the identity mapping loss presented in \cite{zhu2017unpaired}, we also introduce a better identity preserving model that employs the state-of-the-art face verification network (FaceNet) \cite{schroff2015facenet} to ensure the faces are from the same identity. To train the whole translation network, we suggest an identity-aware cyclic GAN loss in an adversarial optimization on the translation between heterogeneous samples. In short, this paper brings two main innovations:
\begin{itemize}
	\item To the best of our knowledge, this paper makes the first attempt to study the problem of unpaired face translation between images and videos.
	
	\item We propose bidirectional image-video translation models with exploiting different losses for the problem of image-to-video and video-to-image face translation. 
\end{itemize}

\section{Background}

\subsection{Image/Video Generation}

As one of the most significant improvements on the research of image generation, Generative Adversarial Networks (GANs) \cite{goodfellow2014generative, radford2015unsupervised, zhao2016energy, berthelot2017began, mao2016least} have drawn substantial attention from both the deep learning and computer vision society. The min-max two-player game involving in GANs provides a simple yet powerful way to estimate target distribution and generate novel image samples. Generally, the framework of GANs consists of a generative model $G$ and a discriminative model $D$, one of which synthesizes highly-realistic images while the other of which distinguishes between synthesized images and real images. The state-of-the-art image GAN technique is Wasserstein GANs \cite{arjovsky2017wasserstein, gulrajani2017improved} which empower the discriminative model with a critic $C$ (replacing the discriminator D) of Earth-move distance (Wasserstein-1) to make the sample distribution close to the real data distribution:
\begin{equation}
\begin{aligned}
\mathcal{L}_{\text{WGAN}} & = \min_{G} \max_{C}  \mathbb{E}_{\bm{x} \sim \mathbb{P}_r} [C(\bm{x})]  -\mathbb{E}_{G(\bm{z}) \sim \mathbb{P}_g} [C(G(\bm{z}))]\\ & + \lambda \mathbb{E}_{\bm{\hat{x}} \sim \mathbb{P}_g} [(\|\nabla_{\bm{\hat{x}}} C(\bm{\hat{x}})\|_2-1)^2],
\end{aligned}
\label{Eq1}
\end{equation}
where $\bm{z}$ is random noise, $\bm{\hat{x}}$ is random samples following the distribution $\mathbb{P}_g$, $G, C$ denotes the mapping functions of generator and critic respectively, and the hyperparameter $\lambda$ controls the balance between the basic GAN loss and the gradient penalty regularization that aims to ensure the mapping function of the critic to be Lipschitz continuous. 

As steady progress has been made toward better image generation, it is also important to study the video generation problem. Although the generated data has only one more dimension over time, the extension from generating images to generating videos typically turns out to be a much more challenging task for the following reasons. First, as a video sequence is a spatio-temporal recording of visual information of objects performing various actions, a generative model on video samples needs to learn the moving patterns of objects in addition to learning their appearance patterns. If the learned object constructs are incorrect, the generated videos typically contain objects performing physically implausible motion. Second, the temporal dimension brings in a huge amount of variation. This would be much more challenging when the background of videos is moving as well. Especially for video-based face generation, as the focus is on faces, one typically uses face detector to crop the involved faces while introducing more unstable changes in both foreground and background. 

To our knowledge, there exist a limited number of video generation techniques \cite{vondrick2016generating, tulyakov2017mocogan} through GANs. In particular, the state-of-the-art method \cite{vondrick2016generating} proposes two models: one employs one-stream video generation network, and the other contains two-stream generation networks, whose first stream serves to generate static background while the second stream produces dynamic foreground. \cite{vondrick2016generating} suggested to use the two-stream model that often outperforms the one-stream model. However, it is intuitive that the two-stream model works well only for videos with static backgrounds with requiring an extra pre-processing step of background stabilization. In fact, the task of background stabilization is typically non-trivial and even may be impossible for videos in the wild. In this paper, we propose to apply the image-based Wasserstein GAN approach to video generation with using one-stream model to overcome the moving background issue that is often happened in video of faces.

\subsection{Image-to-Image Translation}
Among the first attempts, \cite{hertzmann2001image} suggested the problem of image-to-image translation, and employed a non-parametric texture model \cite{efros1999texture} from a single input-output training image pair. Recently, to address the problem, \cite{isola2016image} built a pixel-to-pixel framework by utilizing a conditional generative
adversarial network \cite{goodfellow2014generative} to learn a mapping from input to
output images. However, for many practical tasks, paired training data will not be available. Hence, several other works \cite{liu2017unsupervised, liu2016coupled, rosales2003unsupervised} attempted to tackle the unpaired setting, where the goal is to relate two data domains, $\bm{X}$ and $\bm{Y}$.

More recently, \cite{zhu2017unpaired} proposed Cycle-Consistent Adversarial Netowrks (CycleGAN) to achieve state of the art for the problem of unpaired image-to-image translation. In addition to the standard GAN loss respectively for $\bm{X}$ and $\bm{Y}$, a pair of cycle consistency losses (forward and backward) was formulated using L1 reconstruction loss. For forward cycle consistency, given $\bm{x} \in \bm{X}$ the image translation cycle
should reproduce $\bm{x}$. Backward cycle consistency is similar. Formally, the whole object of CycleGAN is
\begin{equation}
\begin{aligned}
\mathcal{L}(G_{X}, G_{Y}, D_{X}, D_{Y}) & =  \min_{G_X, G_Y} \max_{D_X, D_Y}  \mathcal{L}_{\text{GAN}}(G_{X}, D_{X})\\
& + \mathcal{L}_{\text{GAN}}(G_{Y}, D_{Y}) \\
& + \gamma \mathcal{L}_{\text{cyc}}(G_{X}, G_{Y}),
\end{aligned}
\label{Eq2}
\end{equation}
where $\lambda$ controls the relative importance between the GAN losses $\mathcal{L}_{\text{GAN}}(G_{X}, D_{X})$, $\mathcal{L}_{\text{GAN}}(G_{Y}, D_{Y})$ and the cycle consistency loss $\mathcal{L}_{cyc}(G_{X}, G_{Y})$:
\begin{equation}
\begin{aligned}
\mathcal{L}_{\text{GAN}}(G_{X}, D_{X}) & = \mathbb{E}_{\bm{x} \sim \mathbb{P}_{X_r}} [\log D_X(\bm{x})] \\ & -\mathbb{E}_{G_X(\bm{y}) \sim \mathbb{P}_{X_g}} [\log 
D_X(G_X(\bm{y}))].
\end{aligned}
\label{Eq3}
\end{equation}
\begin{equation}
\begin{aligned}
\mathcal{L}_{\text{GAN}}(G_{Y}, D_{Y}) & =  \mathbb{E}_{\bm{y} \sim \mathbb{P}_{Y_r}} [\log D_Y(\bm{y})] \\ & -\mathbb{E}_{G_Y(\bm{x}) \sim \mathbb{P}_{Y_g}} [\log 
D_Y(G_Y(\bm{x}))].
\end{aligned}
\label{Eq4}
\end{equation}
\begin{equation}
\begin{aligned}
\mathcal{L}_{\text{cyc}}(G_{X}, G_{Y}) & = \mathbb{E}_{\bm{x} \sim \mathbb{P}_{X_g}} [\|G_{X}(G_{Y}(\bm{x})) - \bm{x}\|_1] \\
& + \mathbb{E}_{\bm{y} \sim \mathbb{P}_{\bm{Y}_g}} [\|G_{Y}(G_{X}(\bm{y})) - \bm{y}\|_1].
\end{aligned}
\label{Eq5}
\end{equation}

In this paper, our focus is on image-to-video and video-to-image translation by leveraging the GAN techniques. In contrast to the existing techniques for image-to-image translation, our proposed GAN approach allows hybrid image-based and video-based models to work together which proposes more challenges. In addition, as usual applications on faces, translation on faces is also quite different from translation on general objects.

\subsection{Identity-aware Face Translation}

Face generation can be regarded as a kind of fine-grained object generations due to its special attributes. On one hand, in contrast to general objects, the global structures of faces are approximately the same such that each face can be aligned easier with a universal mean face template. Therefore, one typically can first detect faces in images, and then crop and align the faces for better generation. On the other hand, the variety of real faces are as much as general objects. Accordingly, it is also very challenging to generate identity-preserving faces from static images and even from dynamic videos. 

Preserving facial identity has also been explored in synthesizing the corresponding frontal face image from a single side-view face image \cite{huang2017beyond}, where the identity preserving loss was defined based on the activations of the last two layers of the light CNN \cite{wu2015light}. In multi-view image generation from a single view \cite{zhao2017multi}, a condition image (e.g. frontal view) was used to constrain the generated multiple views in their coarse-to-fine framework.\footnote{There exist other works such as \cite{li2017generate, lu2017conditional} for identity-preserving face generation, most of which however have not been formally published to our best knowledge.} However, such works only designed generative models for image-based face generation rather than video-based generation which is much more challenging as studied before. Hence, to the best of our knowledge, this paper make the first attempt to generate identity-preserving faces in the context of image-to-video and video-to-image translation.

\begin{figure}[t]
	\begin{center}
		\includegraphics[width=1\linewidth]{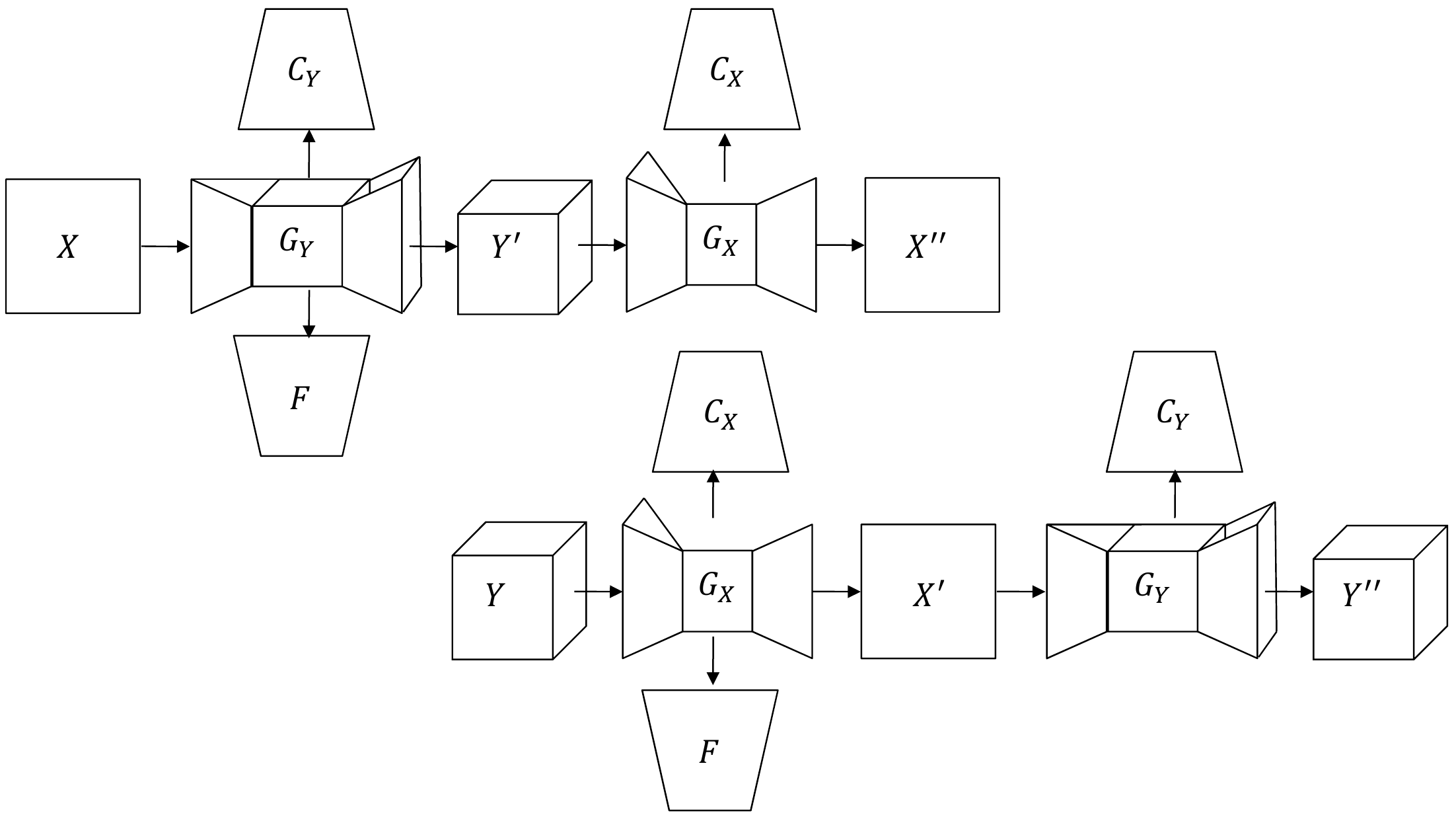}
	\end{center}
	\caption{Conceptual illustration of the proposed Identity-aware CycleGAN (IdCycleGAN). Here $X$, $Y$, $X^{'}$, $Y^{'}$, $X^{''}$, $Y^{''}$ denotes input images, input videos, output images and output videos, reconstructed images and reconstructed videos. $G_{X}, G_{Y}, C_X, C_Y, F$ represents image generator, video generator, image critic, video critic and face verificator respectively.}
	\label{fig:long}
	\label{Fig1}
\end{figure}

\begin{figure}[t]
	\begin{center}
		\includegraphics[width=1\linewidth]{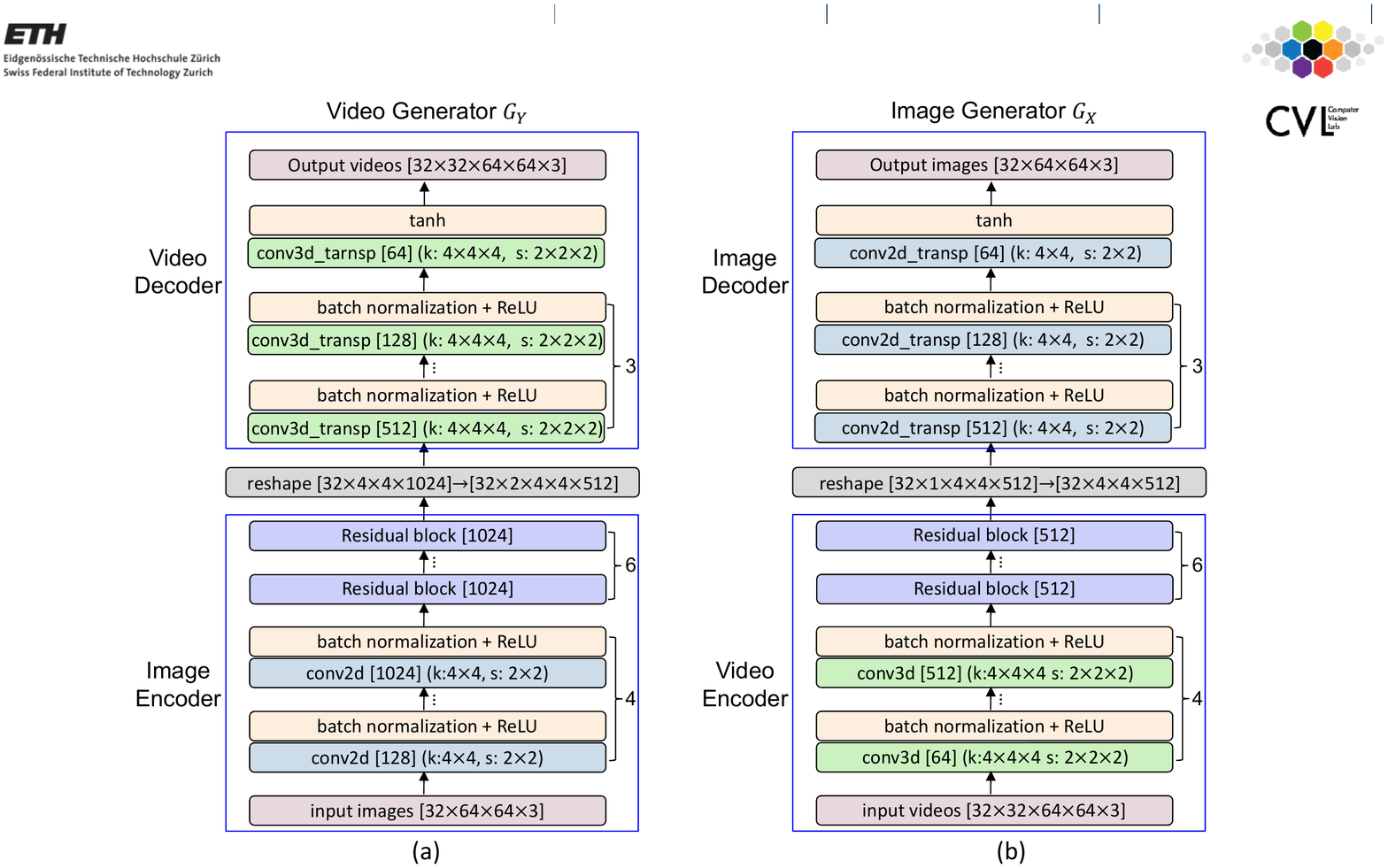}
	\end{center}
	\begin{center}
		\includegraphics[width=0.95\linewidth]{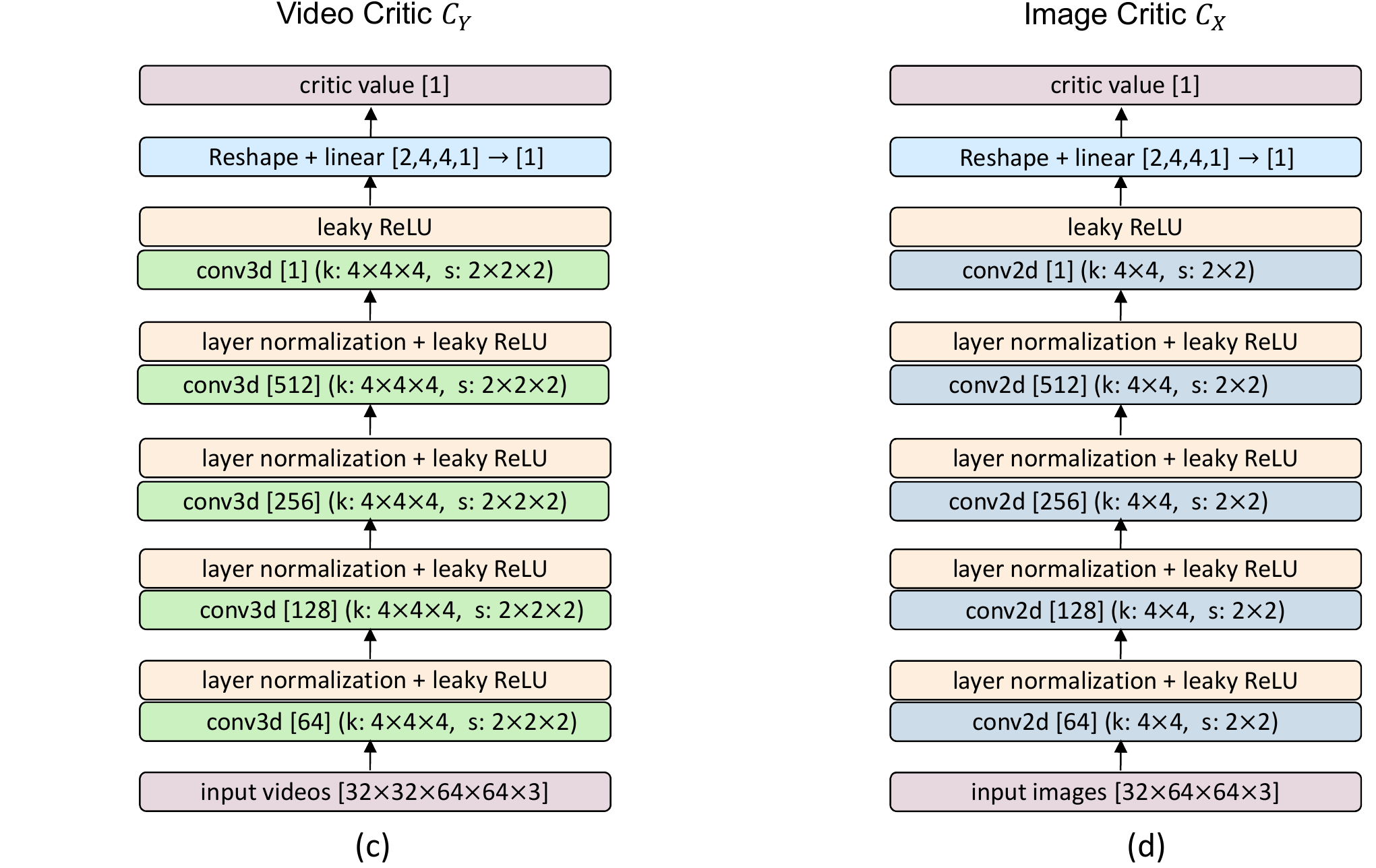}
	\end{center}
	\caption{Architectures of the designed image-to-video translator (video generator), video-to-image translator (image generator), critic networks for images and videos respectively. The digits on the right of the big parantheses indicate the number of repeated components.}
	\label{fig:long}
	\label{Fig2}
\end{figure}

\section{Our Model}

The goal of our method is to learn mapping functions between image domain $\bm{X}$ and video domain $\bm{Y}$ given training image samples $\{\bm{x}_i\}_{i=1}^N \in \bm{X}$ and video samples $\{\bm{y}_i\}_{j=1}^M \in \bm{Y}$. As illustrated in Fig.\ref{Fig1}, our model follows CycleGAN \cite{zhu2017unpaired} to include two inverse mappings: $G_{X}: \bm{X} \rightarrow \bm{Y}$ and $G_{Y}: \bm{Y} \rightarrow \bm{X}$. Following the success of Wasserstein GANs \cite{arjovsky2017wasserstein, gulrajani2017improved}, we also introduce two critic model $C_{\bm{X}}$ and $C_{Y}$, where $C_{X}$ aims to disambiguate between real images ${\bm{x}}$ and translated images ${G_{X}(\bm{y})}$, while $C_{Y}$ aims to distinguish between real videos ${\bm{y}}$ and translated videos ${G_{Y}(\bm{x})}$. In addition, we also introduce a face identification network $F(\bm{x}, \bm{y})$ \cite{schroff2015facenet} to enable  the translated images/videos preserve the identity of the input faces better from input images/videos.

\subsection{Image-Video CycleGAN}

For the image-to-video translation mapping $G_{\bm{Y}}$, our model designs a novel translator network that incorporates image decoder and video encoder; In the same way, for video-to-image translation $G_{\bm{X}}$, we devise a translator network that cascades video decoder with image generator.
As shown in Fig.\ref{Fig2}(a), the image-to-video translator receives a mini-batch of training image samples $\{\bm{x}_i\}_{i=1}^B$ ($B$ is the batch size) as input, then applies an image decoder to decode the images to lower-dimensional latent codes $\{\bm{z}_i\}_{i=1}^B$, followed by a video decoder that finally produces video samples $\{\bm{y}_i\}_{i=1}^B$ containing $l$ frames of size $r \times r$. More specially, the image decoder contains four fractionally-strided 2D convolutional layers, each of which is followed one batch normalization layer and one ReLU layer. Following \cite{zhu2017unpaired}, we also insert three residual blocks into the the third and the fourth 2D convolutional layers respectively. After the image decoding process, we apply a video decoder to produce video samples from the encoded latent codes. The video decoder consists of four layers of fractionally-strided 3D convolutional layers \cite{tran2015learning}, each of which works together with one batch normalization layer and one ReLU layer. The setting has proven to be an efficient approach to upsample while still maintaining spatial and temporal invariances \cite{tran2015learning}.  Similarly, as illustrated in Fig.\ref{Fig2} (b), the video-to-image translator contains a video encoder that encodes the input videos to latent codes, and a image decoder generates images from the resulting latent codes. 

In order to generate realistic images and videos simultaneously, we apply the Wasserstein GAN framework to the setting of image-video translation. In particular, we design two different critic networks respectively for image generation and video generation. As shown in Fig.\ref{Fig2} (c), the image critic network follows the architecture design of \cite{arjovsky2017wasserstein, gulrajani2017improved}, while the video critic model (see Fig.\ref{Fig2} (d)) is specially designed to measure the Wasserstein distance between generated videos and real videos. In particular, the video critic network accepts a video as input, and maps it to a real valued input. It contains five convolutional layers as done in \cite{arjovsky2017wasserstein, gulrajani2017improved} but is then followed by an additional linear layer. Like the image/video generator designed above, we also employ spatio-temporal convolutions with a bias to the output. The convolutions also include a certain of strides in order to allow efficient down-sampling of the high dimensional videos. Following \cite{gulrajani2017improved}, we replace regular batch normalization \cite{ioffe2015batch} layers with layer normalization \cite{ba2016layer} layers as batch normalization is not valid in the setting of penalizing the norm of the critc’s gradient with respect to each input independently. In addition, all but the last convolutional layers use a leaky ReLU activation function. As \cite{arjovsky2017wasserstein, gulrajani2017improved}, we finally use a linear transformation layer to obtain the expectation of the input video samples such that the Wasserstein distance between real and fake samples can be calculated.

With designing such image-video translators and critic models for images and videos, we finally apply Wasserstein loss Eqn.\ref{Eq1} to both the image-to-video mapping $G_{Y}$ and video-to-image mapping $G_{X}$. So far, the network can only map the same set of input images/videos to any random permutation of videos/images in the target domain, where any of the learned mappings can derive an output distribution that approximate the target distribution. Thus, inspired by \cite{zhu2017unpaired}, we additionally leverage the cycle-consistent loss, with which an individual input can be mapped to a desired output by further reducing the space of possible mapping functions. Hence, we express  the objective of our image-to-video/video-to-image translation model as
\begin{equation}
\begin{aligned}
\mathcal{L}(G_{X}, G_{Y}, C_{X}, C_{Y}) & =  \min_{G_X, G_Y} \max_{C_X, C_Y}  \mathcal{L}_{\text{WGAN}}(G_{X}, C_{X})\\
& + \mathcal{L}_{\text{WGAN}}(G_{Y}, C_{Y}) \\
& + \gamma \mathcal{L}_{\text{cyc}}(G_{X}, G_{Y}),
\end{aligned}
\label{Eq6}
\end{equation}
where the involved Wasserestein loss $\mathcal{L}_{\text{WGAN}}(G_{X}, C_{X})$ and $\mathcal{L}_{\text{WGAN}}(G_{Y}, C_{Y})$ respects the form of Eqn.\ref{Eq1}, and the cycle consistency loss $\mathcal{L}_{\text{cyc}}(G_{X}, G_{Y})$ can be achieved by Eqn.\ref{Eq5}. In experiments, we observe that the Wasserstein loss can make the whole training process more stable than using other traditional GAN losses.

\subsection{Identity-aware CycleGAN}

It is still very hard to make the input face and the resulting face being from the same identity. Therefore, we resort to one more constraint that can preserve identity during face translation. One of the most straightforward ways is to apply L2 reconstruction error between the translated samples and the real samples. As studied in \cite{zhu2017unpaired}, to encourage the mapping to preserve color composition between the input and output, it is helpful to introduce an additional loss, i.e., pixel-wise identity mapping loss:
\begin{equation}
\begin{aligned}
\mathcal{L}_{\text{id}}(G_{X}, G_{Y}) & = \mathbb{E}_{\bm{x} \sim \mathbb{P}_{X_g}} [\|(G_{Y}(\bm{x}))_i - \bm{x}\|_2] \\
& + \mathbb{E}_{\bm{y} \sim \mathbb{P}_{Y_g}} [\|G_{X}(\bm{y}) - \bm{y}_i\|_2].
\end{aligned}
\label{Eq7}
\end{equation}

Intuitively, the approach can merely maintain the original appearance on the pixel level. As proved in \cite{schroff2015facenet}, the faithful identity metric can be better achieved on in low-dimensional embedding space. Accordingly, we alternatively come up with using favorable identity metric loss to keep the identity of the original input faces when generating output faces. To this end, we additionally insert a verificator to verify whether the translated face and the input face belong to the same identity. 

In this paper, we exploit the state-of-the-art face verification network (FaceNet) \cite{schroff2015facenet}, which has surpassed human recognition accuracy on some standard face datasets. Essentially, FaceNet maps face images to an embedded space where the squared L2 distance directly correspond to the similarities of the involved faces. As done in the original FaceNet work \cite{schroff2015facenet} that treats video-to-video face verification, we use the average similarity of all pairs of the real (generated) image and the generated (real) frames within the same video clip. The lower score is, the more similar the facial image and video are. Formally, the additional identity-preserving loss is expressed as
\begin{equation}
\begin{aligned}
\mathcal{L}_{\text{id}}(G_{X}, G_{Y}) & = \mathbb{E}_{\bm{x} \sim \mathbb{P}_{X_g}} [\|F((G_{Y}(\bm{x}))_i) - F(\bm{x})\|_2] \\
& + \mathbb{E}_{\bm{y} \sim \mathbb{P}_{Y_g}} [\|F(G_{X}(\bm{y})) - F(\bm{y}_i)\|_2].
\end{aligned}
\label{Eq8}
\end{equation}

\begin{figure*}[t]
	\begin{center}
		\begin{minipage}[t]{0.035\linewidth}
			\centering
			\includegraphics[width=0.267in]{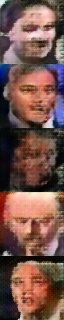}
			\tiny{\text{Frame1}}
			\label{fig:side:b}
		\end{minipage}
		\begin{minipage}[t]{0.035\linewidth}
			\centering
			\includegraphics[width=0.267in]{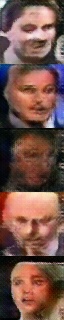}
			\tiny{\text{Frame4}}
			\label{fig:side:b}
		\end{minipage}
		\begin{minipage}[t]{0.035\linewidth}
			\centering
			\includegraphics[width=0.267in]{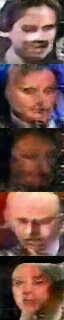}
			\tiny{\text{Frame8}}
			\label{fig:side:b}
		\end{minipage}
		\begin{minipage}[t]{0.035\linewidth}
			\centering
			\includegraphics[width=0.267in]{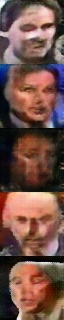}
			\tiny{\text{Frame12}}
			\label{fig:side:b}
		\end{minipage}
		\begin{minipage}[t]{0.035\linewidth}
			\centering
			\includegraphics[width=0.267in]{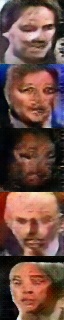}
			\tiny{\text{(d)VideoGAN1}}
			\label{fig:side:b}
		\end{minipage}
		\begin{minipage}[t]{0.035\linewidth}
			\centering
			\includegraphics[width=0.267in]{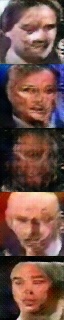}
			\label{fig:side:b}
		\end{minipage}
		\begin{minipage}[t]{0.035\linewidth}
			\centering
			\includegraphics[width=0.267in]{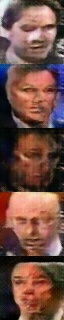}
			\tiny{\text{Frame28}}
			\label{fig:side:b}
		\end{minipage}
		\begin{minipage}[t]{0.04\linewidth}
			\centering
			\includegraphics[width=0.267in]{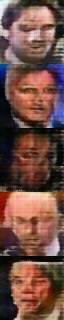}
			\tiny{\text{Frame32}}
			\label{fig:side:b}
		\end{minipage}
		\begin{minipage}[t]{0.035\linewidth}
			\centering
			\includegraphics[width=0.267in]{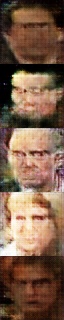}
			\tiny{\text{Frame1}}
			\label{fig:side:b}
		\end{minipage}
		\begin{minipage}[t]{0.035\linewidth}
			\centering
			\includegraphics[width=0.267in]{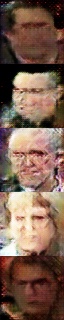}
			\tiny{\text{Frame4}}
			\label{fig:side:b}
		\end{minipage}
		\begin{minipage}[t]{0.035\linewidth}
			\centering
			\includegraphics[width=0.267in]{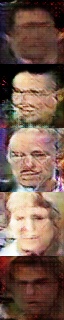}
			\tiny{\text{Frame8}}
			\label{fig:side:b}
		\end{minipage}
		\begin{minipage}[t]{0.035\linewidth}
			\centering
			\includegraphics[width=0.267in]{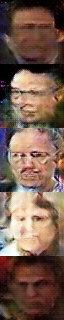}
			\tiny{\text{Frame12}}
			\label{fig:side:b}
		\end{minipage}
		\begin{minipage}[t]{0.035\linewidth}
			\centering
			\includegraphics[width=0.267in]{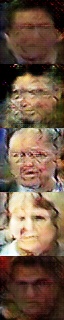}
			\tiny{\text{(b)VideoGAN2}}
			\label{fig:side:b}
		\end{minipage}
		\begin{minipage}[t]{0.035\linewidth}
			\centering
			\includegraphics[width=0.267in]{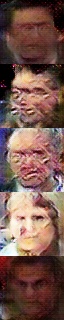}
			\label{fig:side:b}
		\end{minipage}
		\begin{minipage}[t]{0.035\linewidth}
			\centering
			\includegraphics[width=0.267in]{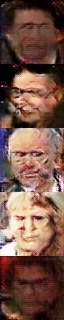}
			\tiny{\text{Frame28}}
			\label{fig:side:b}
		\end{minipage}
		\begin{minipage}[t]{0.04\linewidth}
			\centering
			\includegraphics[width=0.267in]{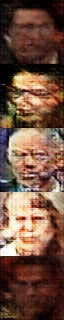}
			\tiny{\text{Frame32}}
			\label{fig:side:b}
		\end{minipage}
		\begin{minipage}[t]{0.035\linewidth}
			\centering
			\includegraphics[width=0.267in]{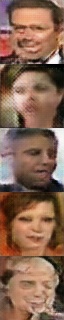}
			\tiny{\text{Frame1}}
			\label{fig:side:b}
		\end{minipage}
		\begin{minipage}[t]{0.035\linewidth}
			\centering
			\includegraphics[width=0.267in]{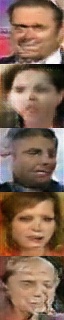}
			\tiny{\text{Frame4}}
			\label{fig:side:b}
		\end{minipage}
		\begin{minipage}[t]{0.035\linewidth}
			\centering
			\includegraphics[width=0.267in]{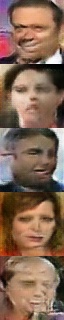}
			\tiny{\text{Frame8}}
			\label{fig:side:b}
		\end{minipage}
		\begin{minipage}[t]{0.035\linewidth}
			\centering
			\includegraphics[width=0.267in]{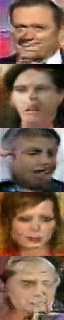}
			\tiny{\text{Frame24}}
			\label{fig:side:b}
		\end{minipage}
		\begin{minipage}[t]{0.035\linewidth}
			\centering
			\includegraphics[width=0.267in]{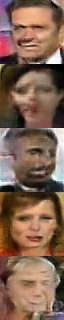}
			\tiny{\text{(c)Ours}}
			\label{fig:side:b}
		\end{minipage}
		\begin{minipage}[t]{0.035\linewidth}
			\centering
			\includegraphics[width=0.267in]{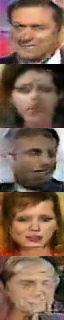}
			\tiny{\text{Frame20}}
			\label{fig:side:b}
		\end{minipage}
		\begin{minipage}[t]{0.035\linewidth}
			\centering
			\includegraphics[width=0.267in]{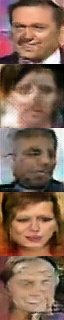}
			\tiny{\text{Frame28}}
			\label{fig:side:b}
		\end{minipage}
		\begin{minipage}[t]{0.035\linewidth}
			\centering
			\includegraphics[width=0.267in]{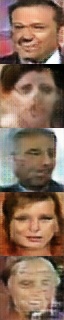}
			\tiny{\text{Frame32}}
			\label{fig:side:b}
		\end{minipage}	
		
		\begin{minipage}[t]{0.035\linewidth}
			\centering
			\includegraphics[width=0.267in]{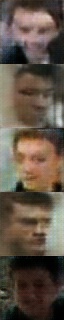}
			\tiny{\text{Frame1}}
			\label{fig:side:b}
		\end{minipage}
		\begin{minipage}[t]{0.035\linewidth}
			\centering
			\includegraphics[width=0.267in]{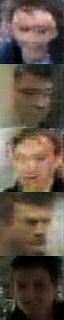}
			\tiny{\text{Frame4}}
			\label{fig:side:b}
		\end{minipage}
		\begin{minipage}[t]{0.035\linewidth}
			\centering
			\includegraphics[width=0.267in]{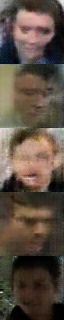}
			\tiny{\text{Frame8}}
			\label{fig:side:b}
		\end{minipage}
		\begin{minipage}[t]{0.035\linewidth}
			\centering
			\includegraphics[width=0.267in]{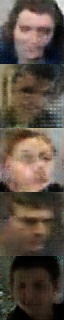}
			\tiny{\text{Frame12}}
			\label{fig:side:b}
		\end{minipage}
		\begin{minipage}[t]{0.035\linewidth}
			\centering
			\includegraphics[width=0.267in]{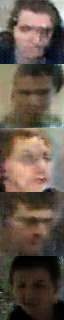}
			\tiny{\text{(d)VideoGAN1}}
			\label{fig:side:b}
		\end{minipage}
		\begin{minipage}[t]{0.035\linewidth}
			\centering
			\includegraphics[width=0.267in]{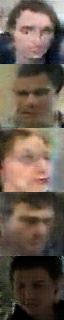}
			\label{fig:side:b}
		\end{minipage}
		\begin{minipage}[t]{0.035\linewidth}
			\centering
			\includegraphics[width=0.267in]{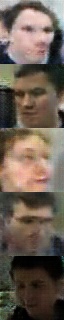}
			\tiny{\text{Frame28}}
			\label{fig:side:b}
		\end{minipage}
		\begin{minipage}[t]{0.04\linewidth}
			\centering
			\includegraphics[width=0.267in]{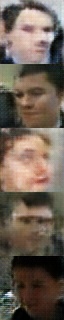}
			\tiny{\text{Frame32}}
			\label{fig:side:b}
		\end{minipage}	
		\begin{minipage}[t]{0.035\linewidth}
			\centering
			\includegraphics[width=0.267in]{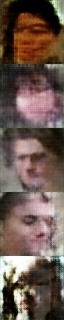}
			\tiny{\text{Frame1}}
			\label{fig:side:b}
		\end{minipage}
		\begin{minipage}[t]{0.035\linewidth}
			\centering
			\includegraphics[width=0.267in]{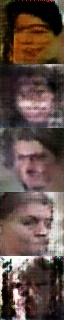}
			\tiny{\text{Frame4}}
			\label{fig:side:b}
		\end{minipage}
		\begin{minipage}[t]{0.035\linewidth}
			\centering
			\includegraphics[width=0.267in]{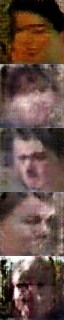}
			\tiny{\text{Frame8}}
			\label{fig:side:b}
		\end{minipage}
		\begin{minipage}[t]{0.035\linewidth}
			\centering
			\includegraphics[width=0.267in]{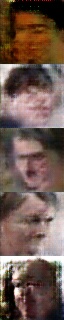}
			\tiny{\text{Frame12}}
			\label{fig:side:b}
		\end{minipage}
		\begin{minipage}[t]{0.035\linewidth}
			\centering
			\includegraphics[width=0.267in]{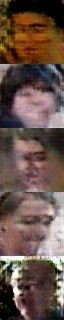}
			\tiny{\text{(e)VideoGAN2}}
			\label{fig:side:b}
		\end{minipage}
		\begin{minipage}[t]{0.035\linewidth}
			\centering
			\includegraphics[width=0.267in]{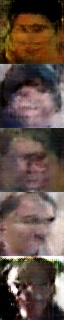}
			\label{fig:side:b}
		\end{minipage}
		\begin{minipage}[t]{0.035\linewidth}
			\centering
			\includegraphics[width=0.267in]{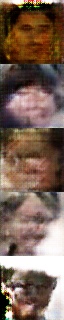}
			\tiny{\text{Frame28}}
			\label{fig:side:b}
		\end{minipage}
		\begin{minipage}[t]{0.04\linewidth}
			\centering
			\includegraphics[width=0.267in]{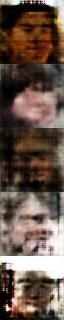}
			\tiny{\text{Frame32}}
			\label{fig:side:b}
		\end{minipage}
		\begin{minipage}[t]{0.035\linewidth}
			\centering
			\includegraphics[width=0.267in]{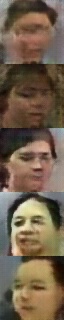}
			\tiny{\text{Frame1}}
			\label{fig:side:b}
		\end{minipage}
		\begin{minipage}[t]{0.035\linewidth}
			\centering
			\includegraphics[width=0.267in]{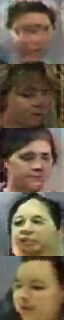}
			\tiny{\text{Frame4}}
			\label{fig:side:b}
		\end{minipage}
		\begin{minipage}[t]{0.035\linewidth}
			\centering
			\includegraphics[width=0.267in]{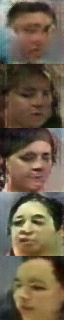}
			\tiny{\text{Frame8}}
			\label{fig:side:b}
		\end{minipage}
		\begin{minipage}[t]{0.035\linewidth}
			\centering
			\includegraphics[width=0.267in]{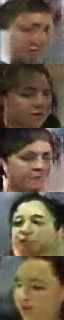}
			\tiny{\text{Frame12}}
			\label{fig:side:b}
		\end{minipage}
		\begin{minipage}[t]{0.035\linewidth}
			\centering
			\includegraphics[width=0.267in]{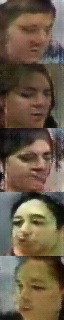}
			\tiny{\text{(f)Ours}}
			\label{fig:side:b}
		\end{minipage}
		\begin{minipage}[t]{0.035\linewidth}
			\centering
			\includegraphics[width=0.267in]{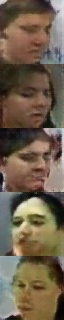}
			\label{fig:side:b}
		\end{minipage}
		\begin{minipage}[t]{0.035\linewidth}
			\centering
			\includegraphics[width=0.267in]{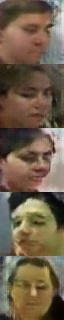}
			\tiny{\text{Frame28}}
			\label{fig:side:b}
		\end{minipage}
		\begin{minipage}[t]{0.04\linewidth}
			\centering
			\includegraphics[width=0.267in]{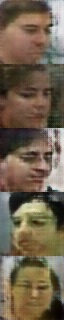}
			\tiny{\text{Frame32}}
			\label{fig:side:b}
		\end{minipage}
		
		\caption{Video generation results of VideoGAN1 (one-stream) \cite{vondrick2016generating}, VideoGAN2 (two-stream) \cite{vondrick2016generating} and our proposed model respectively on the Celeb (a)-(b)-(c) and the PaSC (d)-(e)-(f)}
		\label{fig:generation}  
	\end{center}
\end{figure*}

Consequently, the final objective of the proposed model includes three terms: an adversarial Wasserstein loss $\mathcal{L}_{\text{WGAN}}(G_{X}, C_{X}), \mathcal{L}_{\text{WGAN}}(G_{Y}, C_{Y})$ for matching the distribution of generated images/videos to the data distribution in the target images/videos; a cycle consistency loss $\mathcal{L}_{\text{cyc}}(G_{X}, G_{Y})$ to prevent the learned mappings $G_{X}$ and $G_{Y}$ from contradicting each other; and a identity-preserving loss $\mathcal{L}_{\text{id}}(G_{X}, G_{Y})$ with a hyperparameter $\omega$ to measure the identity metric between the generated images/videos and the input source images/videos.

\section{Experiment}

To evaluate our model for face translation between images and videos, we use standard face datasets: Celebrity Face (combining CelebA \cite{liu2015faceattributes} with YouTubeFace \cite{wolf2011face}) and Point-and-Shoot Challenge (PaSC) \cite{beveridge2013challenge}. 
To justify the effectiveness of our designed image-video translator architecture for robust video generation, we first compare our model against the state-of-the-art video generation technique VideoGAN \cite{vondrick2016generating}. Then we evaluate our proposed image-video translation model with using the Wasserstein setting of the baseline CycleGAN \cite{zhu2017unpaired} loss Eqn.\ref{Eq6} (denoted as \emph{Model1}), adding pixel-wise identity loss Eqn.\ref{Eq7}  (denoted as  \emph{Model2}), or adding face identity loss Eqn.\ref{Eq8} (denoted as \emph{Model3}) on the two used face datasets for face translation between static images and dynamic videos.

\subsection{Experimental Setting}

The used Celebrity Face dataset (Celeb) consists of CelebA \cite{liu2015faceattributes} and YouTubeFace \cite{wolf2011face}, which contains facial images and videos respectively. The CelebA contains 202,599 images in total, and we use the frontal face detector with Haar cascades to get 156,254 frontal faces in total, which can constitute relatively high quality frontal facial images. The YouTubeFace has 3,425 videos of 1,595 different celebrities. An average of 2.15 videos are available for each subject, and the average length of a video clip is 181.3 frames. The video frames often exhibit a large variation of pose and illumination, as well as degradations such as compression effects. We employ the state-of-the-art face detector from \cite{zhang2016joint} to detect faces from the YouTube videos, and finally obtain 625,988 facial frames of 626,146 in total. In contrast, the PaSC dataset \cite{beveridge2013challenge} is designed to simulate more of a video surveillance system. The PaSC was collected for face recognition from stills and videos captured by point-and-shoot cameras. It includes 9,376 still images and 2,802 videos (303,222 frames) of 293 people in total. The data has large variation in terms of distance to the camera, alternative sensors, frontal versus not-frontal views, and varying locations. As done for Celeb, we finally get 9,249 still facial images and 300,206 farcical video frames. Considering the balance of performance and computational cost of the evaluated models, all the faces are center cropped to the size of $64 \times 64$.  

For our proposed IdCycleGAN models (Model1, Model2, Model3), following \cite{gulrajani2017improved}, we set the balance parameter $\lambda$ of the gradient norm penalty regularization to 10. The parameter $\gamma$ of the cycle consistency loss and is fixed as 1000, while the parameter $\omega$ of the identity-aware loss is set to 100. We follow \cite{schroff2015facenet} to employ the inception-ResNet-v1 model to achieve the verifcator model $F$ of our IdCycleGAN, that actually uses the pretrained FaceNet. For the baseline video generation method VideoGAN \cite{vondrick2016generating}, we set the trade-off parameter of using mask to combine the forground and background stream in the context of its two-stream model.
For the baseline image-to-image translation model CycleGAN \cite{zhu2017unpaired}, we adapt it to the scenario of image-video face translation with using our proposed image-video translation model, more stable Wasserstein GAN loss and both forward and backward cycle consistency loss. The training of each deep model was performed on a GeForce GTX TITAN X GPU.

\begin{figure*}[t]
	\begin{center}
		\begin{minipage}[t]{0.08\linewidth}
			\centering
			\includegraphics[width=0.267in]{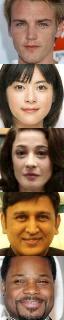}
			\scriptsize{\text{(a)Input Image}}
			\label{fig:side:b}
		\end{minipage}
		\begin{minipage}[t]{0.035\linewidth}
			\centering
			\includegraphics[width=0.267in]{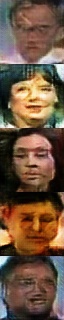}
			\tiny{\text{Frame1}}
			\label{fig:side:b}
		\end{minipage}
		\begin{minipage}[t]{0.035\linewidth}
			\centering
			\includegraphics[width=0.267in]{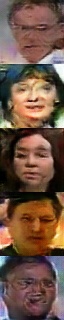}
			\tiny{\text{Frame8}}
			\label{fig:side:b}
		\end{minipage}
		\begin{minipage}[t]{0.035\linewidth}
			\centering
			\includegraphics[width=0.267in]{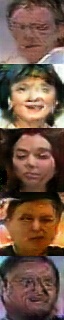}
			\scriptsize{\text{(b)M1}}
			\label{fig:side:b}
		\end{minipage}
		\begin{minipage}[t]{0.035\linewidth}
			\centering
			\includegraphics[width=0.267in]{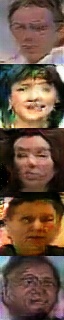}
			\label{fig:side:b}
		\end{minipage}
		\begin{minipage}[t]{0.04\linewidth}
			\centering
			\includegraphics[width=0.267in]{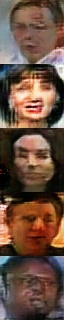}
			\tiny{\text{Frame32}}
			\label{fig:side:b}
		\end{minipage}
		\begin{minipage}[t]{0.08\linewidth}
			\centering
			\includegraphics[width=0.267in]{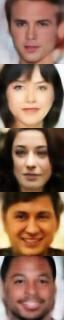}
			\scriptsize{\text{(c)M1-recon}}
			\label{fig:side:b}
		\end{minipage}
		\begin{minipage}[t]{0.035\linewidth}
			\centering
			\includegraphics[width=0.267in]{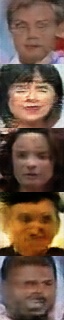}
			\tiny{\text{Frame1}}
			\label{fig:side:b}
		\end{minipage}
		\begin{minipage}[t]{0.035\linewidth}
			\centering
			\includegraphics[width=0.267in]{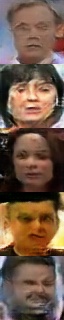}
			\tiny{\text{Frame8}}
			\label{fig:side:b}
		\end{minipage}
		\begin{minipage}[t]{0.035\linewidth}
			\centering
			\includegraphics[width=0.267in]{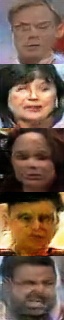}
			\scriptsize{\text{(d)M2}}
			\label{fig:side:b}
		\end{minipage}
		\begin{minipage}[t]{0.035\linewidth}
			\centering
			\includegraphics[width=0.267in]{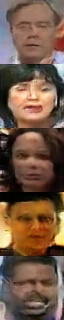}
			\label{fig:side:b}
		\end{minipage}
		\begin{minipage}[t]{0.04\linewidth}
			\centering
			\includegraphics[width=0.267in]{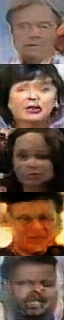}
			\tiny{\text{Frame32}}
			\label{fig:side:b}
		\end{minipage}
		\begin{minipage}[t]{0.08\linewidth}
			\centering
			\includegraphics[width=0.267in]{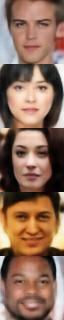}
			\scriptsize{\text{(e)M2-recon}}
			\label{fig:side:b}
		\end{minipage}
		\begin{minipage}[t]{0.035\linewidth}
			\centering
			\includegraphics[width=0.267in]{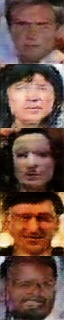}
			\tiny{\text{Frame1}}
			\label{fig:side:b}
		\end{minipage}
		\begin{minipage}[t]{0.035\linewidth}
			\centering
			\includegraphics[width=0.267in]{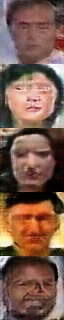}
			\tiny{\text{Frame8}}
			\label{fig:side:b}
		\end{minipage}
		\begin{minipage}[t]{0.035\linewidth}
			\centering
			\includegraphics[width=0.267in]{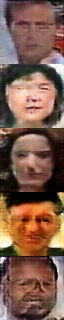}
			\scriptsize{\text{(f)M3}}
			\label{fig:side:b}
		\end{minipage}
		\begin{minipage}[t]{0.035\linewidth}
			\centering
			\includegraphics[width=0.267in]{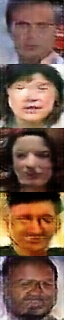}
			\label{fig:side:b}
		\end{minipage}
		\begin{minipage}[t]{0.035\linewidth}
			\centering
			\includegraphics[width=0.267in]{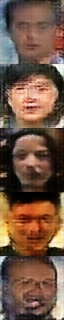}
			\tiny{\text{Frame32}}
			\label{fig:side:b}
		\end{minipage}		
		\begin{minipage}[t]{0.08\linewidth}
			\centering
			\includegraphics[width=0.267in]{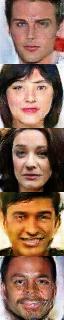}
			\scriptsize{\text{(g)M3-recon}}
			\label{fig:side:b}
		\end{minipage}
		
		\begin{minipage}[t]{0.08\linewidth}
			\centering
			\includegraphics[width=0.267in]{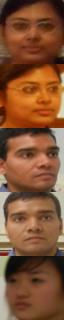}
			\scriptsize{\text{(a)Input Image}}
			\label{fig:side:b}
		\end{minipage}
		\begin{minipage}[t]{0.035\linewidth}
			\centering
			\includegraphics[width=0.267in]{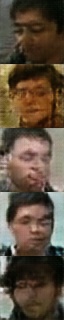}
			\tiny{\text{Frame1}}
			\label{fig:side:b}
		\end{minipage}
		\begin{minipage}[t]{0.035\linewidth}
			\centering
			\includegraphics[width=0.267in]{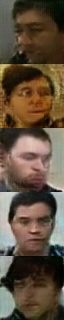}
			\tiny{\text{Frame8}}
			\label{fig:side:b}
		\end{minipage}
		\begin{minipage}[t]{0.035\linewidth}
			\centering
			\includegraphics[width=0.267in]{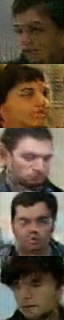}
			\scriptsize{\text{(b)M1}}
			\label{fig:side:b}
		\end{minipage}
		\begin{minipage}[t]{0.035\linewidth}
			\centering
			\includegraphics[width=0.267in]{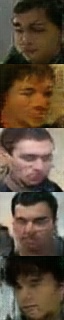}
			\label{fig:side:b}
		\end{minipage}
		\begin{minipage}[t]{0.04\linewidth}
			\centering
			\includegraphics[width=0.267in]{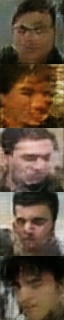}
			\tiny{\text{Frame32}}
			\label{fig:side:b}
		\end{minipage}
		\begin{minipage}[t]{0.08\linewidth}
			\centering
			\includegraphics[width=0.267in]{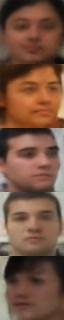}
			\scriptsize{\text{(c)M1-recon}}
			\label{fig:side:b}
		\end{minipage}
		\begin{minipage}[t]{0.035\linewidth}
			\centering
			\includegraphics[width=0.267in]{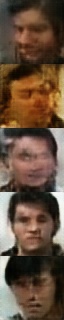}
			\tiny{\text{Frame1}}
			\label{fig:side:b}
		\end{minipage}
		\begin{minipage}[t]{0.035\linewidth}
			\centering
			\includegraphics[width=0.267in]{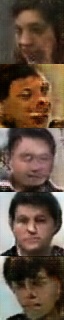}
			\tiny{\text{Frame8}}
			\label{fig:side:b}
		\end{minipage}
		\begin{minipage}[t]{0.035\linewidth}
			\centering
			\includegraphics[width=0.267in]{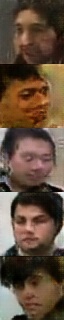}
			\scriptsize{\text{(d)M2}}
			\label{fig:side:b}
		\end{minipage}
		\begin{minipage}[t]{0.035\linewidth}
			\centering
			\includegraphics[width=0.267in]{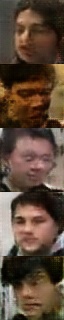}
			\label{fig:side:b}
		\end{minipage}
		\begin{minipage}[t]{0.04\linewidth}
			\centering
			\includegraphics[width=0.267in]{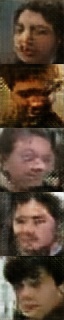}
			\tiny{\text{Frame32}}
			\label{fig:side:b}
		\end{minipage}
		\begin{minipage}[t]{0.08\linewidth}
			\centering
			\includegraphics[width=0.267in]{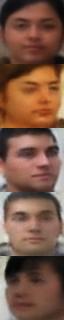}
			\scriptsize{\text{(d)M2-recon}}
			\label{fig:side:b}
		\end{minipage}
		\begin{minipage}[t]{0.035\linewidth}
			\centering
			\includegraphics[width=0.267in]{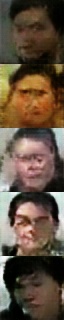}
			\tiny{\text{Frame1}}
			\label{fig:side:b}
		\end{minipage}
		\begin{minipage}[t]{0.035\linewidth}
			\centering
			\includegraphics[width=0.267in]{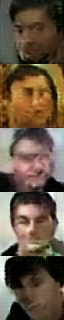}
			\tiny{\text{Frame8}}
			\label{fig:side:b}
		\end{minipage}
		\begin{minipage}[t]{0.035\linewidth}
			\centering
			\includegraphics[width=0.267in]{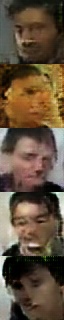}
			\scriptsize{\text{(f)M3}}
			\label{fig:side:b}
		\end{minipage}
		\begin{minipage}[t]{0.035\linewidth}
			\centering
			\includegraphics[width=0.267in]{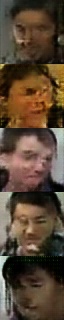}
			\label{fig:side:b}
		\end{minipage}
		\begin{minipage}[t]{0.035\linewidth}
			\centering
			\includegraphics[width=0.267in]{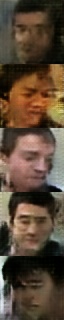}
			\tiny{\text{Frame32}}
			\label{fig:side:b}
		\end{minipage}	
		\begin{minipage}[t]{0.08\linewidth}
			\centering
			\includegraphics[width=0.267in]{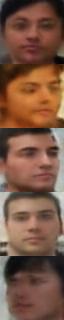}
			\scriptsize{\text{(g)M3-recon}}
			\label{fig:side:b}
		\end{minipage}
		
		\caption{Image-to-Video translation results (b)Model1-(d)Model2-(f)Model3 and their reconstructed results (c)-(e)-(g) on the Celeb and PaSC databases}
		\label{fig:translation_celeb_img2vid}  
	\end{center}
\end{figure*}

\begin{figure*}[t]
	\begin{center}
		\begin{minipage}[t]{0.035\linewidth}
			\centering
			\includegraphics[width=0.267in]{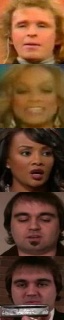}
			\tiny{\text{Frame1}}
			\label{fig:side:b}
		\end{minipage}
		\begin{minipage}[t]{0.035\linewidth}
			\centering
			\includegraphics[width=0.267in]{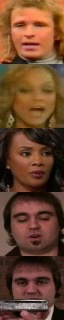}
			\tiny{\text{Frame8}}
			\label{fig:side:b}
		\end{minipage}
		\begin{minipage}[t]{0.035\linewidth}
			\centering
			\includegraphics[width=0.267in]{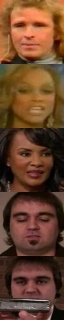}
			\scriptsize{\text{(a)Input Video}}
			\label{fig:side:b}
		\end{minipage}
		\begin{minipage}[t]{0.035\linewidth}
			\centering
			\includegraphics[width=0.267in]{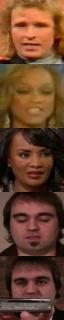}
			\label{fig:side:b}
		\end{minipage}
		\begin{minipage}[t]{0.04\linewidth}
			\centering
			\includegraphics[width=0.267in]{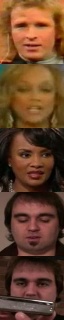}
			\tiny{\text{Frame32}}
			\label{fig:side:b}
		\end{minipage}		
		\begin{minipage}[t]{0.04\linewidth}
			\centering
			\includegraphics[width=0.267in]{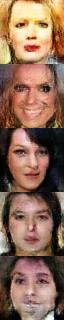}
			\scriptsize{\text{(b)M1}}
			\label{fig:side:b}
		\end{minipage}
		\begin{minipage}[t]{0.035\linewidth}
			\centering
			\includegraphics[width=0.267in]{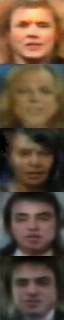}
			\tiny{\text{Frame1}}
			\label{fig:side:b}
		\end{minipage}
		\begin{minipage}[t]{0.035\linewidth}
			\centering
			\includegraphics[width=0.267in]{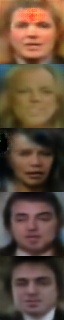}
			\tiny{\text{Frame8}}
			\label{fig:side:b}
		\end{minipage}
		\begin{minipage}[t]{0.035\linewidth}
			\centering
			\includegraphics[width=0.267in]{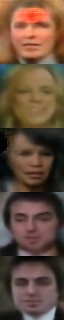}
			\scriptsize{\text{(c)M1-recon}}
			\label{fig:side:b}
		\end{minipage}
		\begin{minipage}[t]{0.035\linewidth}
			\centering
			\includegraphics[width=0.267in]{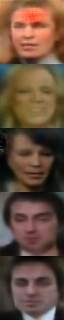}
			\label{fig:side:b}
		\end{minipage}
		\begin{minipage}[t]{0.04\linewidth}
			\centering
			\includegraphics[width=0.267in]{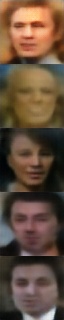}
			\tiny{\text{Frame32}}
			\label{fig:side:b}
		\end{minipage}
		\begin{minipage}[t]{0.04\linewidth}
			\centering
			\includegraphics[width=0.267in]{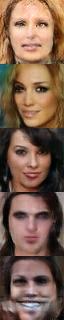}
			\scriptsize{\text{(d)M2}}
			\label{fig:side:b}
		\end{minipage}
		\begin{minipage}[t]{0.035\linewidth}
			\centering
			\includegraphics[width=0.267in]{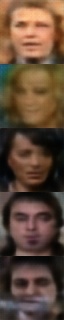}
			\tiny{\text{Frame1}}
			\label{fig:side:b}
		\end{minipage}
		\begin{minipage}[t]{0.035\linewidth}
			\centering
			\includegraphics[width=0.267in]{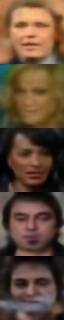}
			\tiny{\text{Frame8}}
			\label{fig:side:b}
		\end{minipage}
		\begin{minipage}[t]{0.035\linewidth}
			\centering
			\includegraphics[width=0.267in]{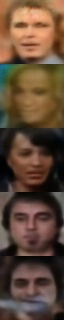}
			\scriptsize{\text{(e)M2-recon}}
			\label{fig:side:b}
		\end{minipage}
		\begin{minipage}[t]{0.035\linewidth}
			\centering
			\includegraphics[width=0.267in]{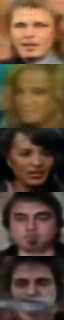}
			\label{fig:side:b}
		\end{minipage}
		\begin{minipage}[t]{0.04\linewidth}
			\centering
			\includegraphics[width=0.267in]{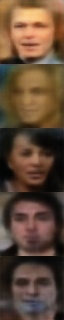}
			\tiny{\text{Frame32}}
			\label{fig:side:b}
		\end{minipage}
		\begin{minipage}[t]{0.04\linewidth}
			\centering
			\includegraphics[width=0.267in]{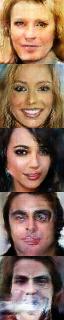}
			\scriptsize{\text{(f)M3}}
			\label{fig:side:b}
		\end{minipage}
		\begin{minipage}[t]{0.035\linewidth}
			\centering
			\includegraphics[width=0.267in]{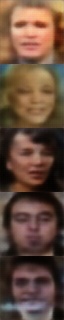}
			\tiny{\text{Frame1}}
			\label{fig:side:b}
		\end{minipage}
		\begin{minipage}[t]{0.035\linewidth}
			\centering
			\includegraphics[width=0.267in]{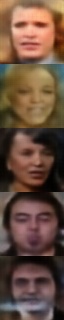}
			\tiny{\text{Frame8}}
			\label{fig:side:b}
		\end{minipage}
		\begin{minipage}[t]{0.035\linewidth}
			\centering
			\includegraphics[width=0.267in]{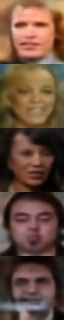}
			\scriptsize{\text{(g)M3-recon}}
			\label{fig:side:b}
		\end{minipage}
		\begin{minipage}[t]{0.035\linewidth}
			\centering
			\includegraphics[width=0.267in]{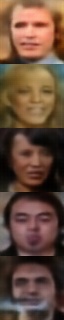}
			\label{fig:side:b}
		\end{minipage}
		\begin{minipage}[t]{0.035\linewidth}
			\centering
			\includegraphics[width=0.267in]{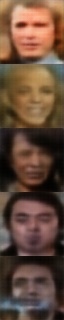}
			\tiny{\text{Frame32}}
			\label{fig:side:b}
		\end{minipage}		
		
		\begin{minipage}[t]{0.035\linewidth}
			\centering
			\includegraphics[width=0.267in]{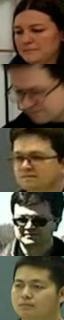}
			\tiny{\text{Frame1}}
			\label{fig:side:b}
		\end{minipage}
		\begin{minipage}[t]{0.035\linewidth}
			\centering
			\includegraphics[width=0.267in]{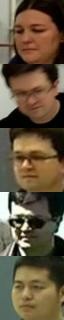}
			\tiny{\text{Frame8}}
			\label{fig:side:b}
		\end{minipage}
		\begin{minipage}[t]{0.035\linewidth}
			\centering
			\includegraphics[width=0.267in]{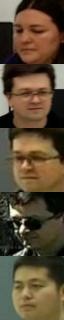}
			\scriptsize{\text{(a)Input Video}}
			\label{fig:side:b}
		\end{minipage}
		\begin{minipage}[t]{0.035\linewidth}
			\centering
			\includegraphics[width=0.267in]{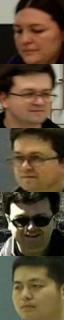}
			\label{fig:side:b}
		\end{minipage}
		\begin{minipage}[t]{0.04\linewidth}
			\centering
			\includegraphics[width=0.267in]{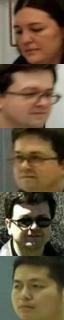}
			\tiny{\text{Frame32}}
			\label{fig:side:b}
		\end{minipage}		
		\begin{minipage}[t]{0.04\linewidth}
			\centering
			\includegraphics[width=0.267in]{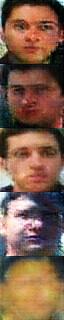}
			\scriptsize{\text{(b)M1}}
			\label{fig:side:b}
		\end{minipage}
		\begin{minipage}[t]{0.035\linewidth}
			\centering
			\includegraphics[width=0.267in]{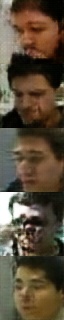}
			\tiny{\text{Frame1}}
			\label{fig:side:b}
		\end{minipage}
		\begin{minipage}[t]{0.035\linewidth}
			\centering
			\includegraphics[width=0.267in]{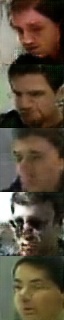}
			\tiny{\text{Frame8}}
			\label{fig:side:b}
		\end{minipage}
		\begin{minipage}[t]{0.035\linewidth}
			\centering
			\includegraphics[width=0.267in]{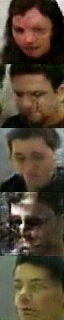}
			\scriptsize{\text{(c)M1-recon}}
			\label{fig:side:b}
		\end{minipage}
		\begin{minipage}[t]{0.035\linewidth}
			\centering
			\includegraphics[width=0.267in]{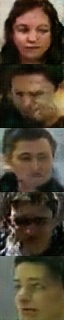}
			\label{fig:side:b}
		\end{minipage}
		\begin{minipage}[t]{0.04\linewidth}
			\centering
			\includegraphics[width=0.267in]{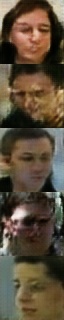}
			\tiny{\text{Frame32}}
			\label{fig:side:b}
		\end{minipage}
		\begin{minipage}[t]{0.04\linewidth}
			\centering
			\includegraphics[width=0.267in]{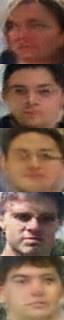}
			\scriptsize{\text{(d)M2}}
			\label{fig:side:b}
		\end{minipage}
		\begin{minipage}[t]{0.035\linewidth}
			\centering
			\includegraphics[width=0.267in]{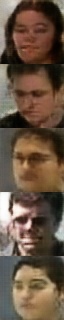}
			\tiny{\text{Frame1}}
			\label{fig:side:b}
		\end{minipage}
		\begin{minipage}[t]{0.035\linewidth}
			\centering
			\includegraphics[width=0.267in]{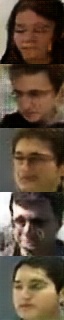}
			\tiny{\text{Frame8}}
			\label{fig:side:b}
		\end{minipage}
		\begin{minipage}[t]{0.035\linewidth}
			\centering
			\includegraphics[width=0.267in]{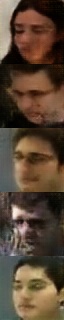}
			\scriptsize{\text{(e)M2-recon}}
			\label{fig:side:b}
		\end{minipage}
		\begin{minipage}[t]{0.035\linewidth}
			\centering
			\includegraphics[width=0.267in]{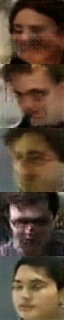}
			\label{fig:side:b}
		\end{minipage}
		\begin{minipage}[t]{0.04\linewidth}
			\centering
			\includegraphics[width=0.267in]{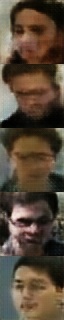}
			\tiny{\text{Frame32}}
			\label{fig:side:b}
		\end{minipage}
		\begin{minipage}[t]{0.04\linewidth}
			\centering
			\includegraphics[width=0.267in]{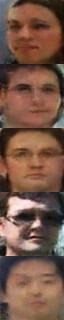}
			\scriptsize{\text{(f)M3}}
			\label{fig:side:b}
		\end{minipage}
		\begin{minipage}[t]{0.035\linewidth}
			\centering
			\includegraphics[width=0.267in]{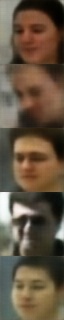}
			\tiny{\text{Frame1}}
			\label{fig:side:b}
		\end{minipage}
		\begin{minipage}[t]{0.035\linewidth}
			\centering
			\includegraphics[width=0.267in]{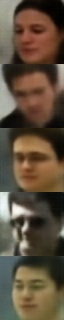}
			\tiny{\text{Frame8}}
			\label{fig:side:b}
		\end{minipage}
		\begin{minipage}[t]{0.035\linewidth}
			\centering
			\includegraphics[width=0.267in]{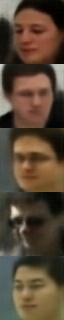}
			\scriptsize{\text{(g)M3-recon}}
			\label{fig:side:b}
		\end{minipage}
		\begin{minipage}[t]{0.035\linewidth}
			\centering
			\includegraphics[width=0.267in]{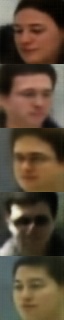}
			\label{fig:side:b}
		\end{minipage}
		\begin{minipage}[t]{0.035\linewidth}
			\centering
			\includegraphics[width=0.267in]{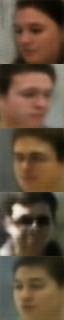}
			\tiny{\text{Frame32}}
			\label{fig:side:b}
		\end{minipage}	
		
		\caption{Video-to-Image translation results (b)Model1-(d)Model2-(f)Model3 and their reconstructed results (c)-(e)-(g) on the Celeb and PaSC databases}
		\label{fig:translation_celeb_vid2img}  
	\end{center}
\end{figure*}

\subsection{Qualitative Evaluation}

\textbf{Video Face Generation}: We first degrade our model of image-video translation to video generation model with merely removing the image decoder out of the video generator and only using the Wasserstein GAN loss for generating videos. In the case, we can evaluate how robust our proposed model performs by comparing with the state-of-the-art video generation model VideoGAN \cite{vondrick2016generating} including its one-stream model and two-stream model. In Fig.\ref{fig:generation}, we show some examples of the videos generated from the VideoGAN one-stream model, the VideoGAN two-stream model and our one-stream model. From the results on the Celeb and PaSC datasets, we can observe that the VideoGAN two-stream model performs badly on both the two datasets whereas the VideoGAN one-stream model works relatively normally. In contrast, our proposed model can perform the best with generating relatively sharper and more realistic faces. This not only verifies that our proposed approach with using one-stream model can perform well in the real-world setting, but also demonstrate that our generalized Wasserstein GAN framework in the context of videos can converge more well in comparison to the state-of-the-art VideoGAN that is under the traditional DCGAN \cite{radford2015unsupervised} framework.

\textbf{Image-Video Face Translation}: Fig.\ref{fig:translation_celeb_img2vid} and Fig.\ref{fig:translation_celeb_vid2img} quantitatively exhibit the results of of the comparing our three models Model1, Model2, Model3 for face translation between images and videos on the Celeb and the PaSC database. Note that we also show some more translated static and dynamic facial samples in Fig.\ref{fig:translation_avi} which is best viewed with the Adobe reader.

In the qualitative evaluation for image-to-video translation  Fig.\ref{fig:translation_celeb_img2vid} (a)-(b)-(d)-(f), we can discover that all the three models can generate visually plausible and temporally continuous facial video frames from input facial images. In addition to using the stable Wasserstein GAN loss, we attribute the good performance to using the proposed image-video generator network that enables the hybrid decoding and encoding process on involved images and videos to work well together. In addition, by comparing with Model1 and Model2, we can find that Model3 performs the best in terms of maintaining the identity of input faces from images/videos. This demonstrates that the proposed identity-aware loss can enhance the capability of preserving identity when translating faces between images and videos.

In the video-to-image translation experiment (Fig.\ref{fig:translation_celeb_vid2img} (a)-(b)-(d)-(f)), on the Celeb face dataset, Model1 can often translate plausible faces between images and videos, but the identity information is not kept during translation like the fourth and the fifth samples. While Model2 can address the problem to some extent, it still fails in some cases like the 5-th facial sample whose mouth is occluded by a melodica. Thanks to imposing the facial identity loss, Model3 can maintain the identity better in the process of face translation. Similar observation can be achieved on the PaSC face dataset by comparing the translated facial samples especially the first and the fourth samples that are shown in (a)-(b)-(d)-(f).

As we can see from some failure cases shown in Fig.\ref{fig:translation_avi}, the non-frontal and quick moving faces are the most difficult to translate for all of the three proposed models. In addition, the visual quality of the Model3's generated samples are not always the best which suggests the identity preserving term would be very strong to prevent the model from getting more visually plausible results.

\begin{figure*}[t]
	\begin{center}
		\begin{minipage}[t]{0.49\linewidth}
			\centering
			\includegraphics[width=1\linewidth]{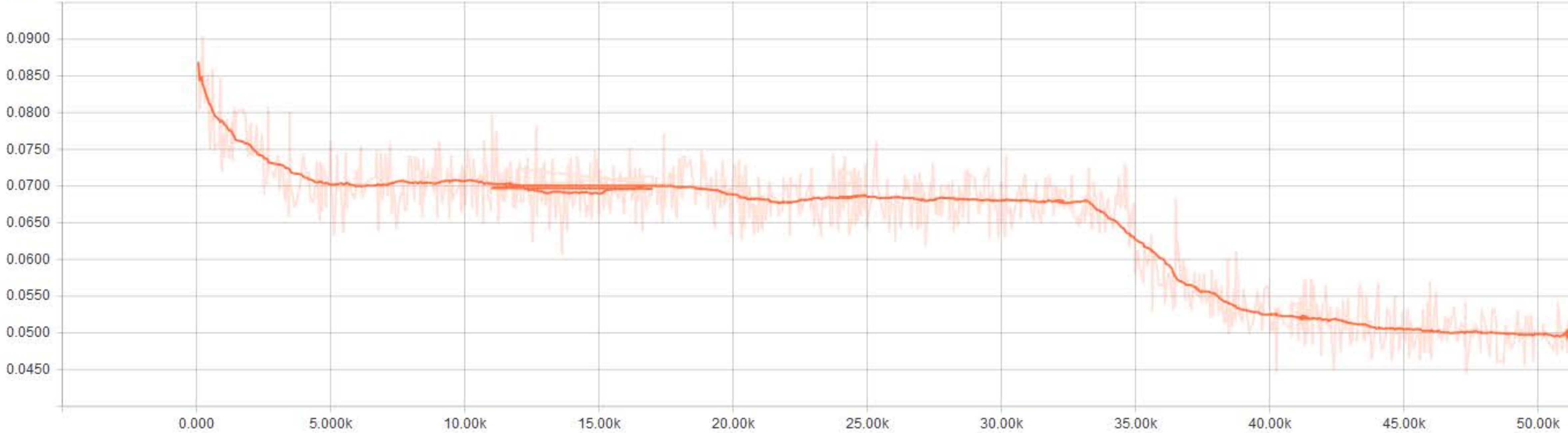}
			\small{\text{(a) Image-to-Video translation on Celeb}}
			\label{fig:side:b}
		\end{minipage}	
		\begin{minipage}[t]{0.49\linewidth}
			\centering
			\includegraphics[width=1\linewidth]{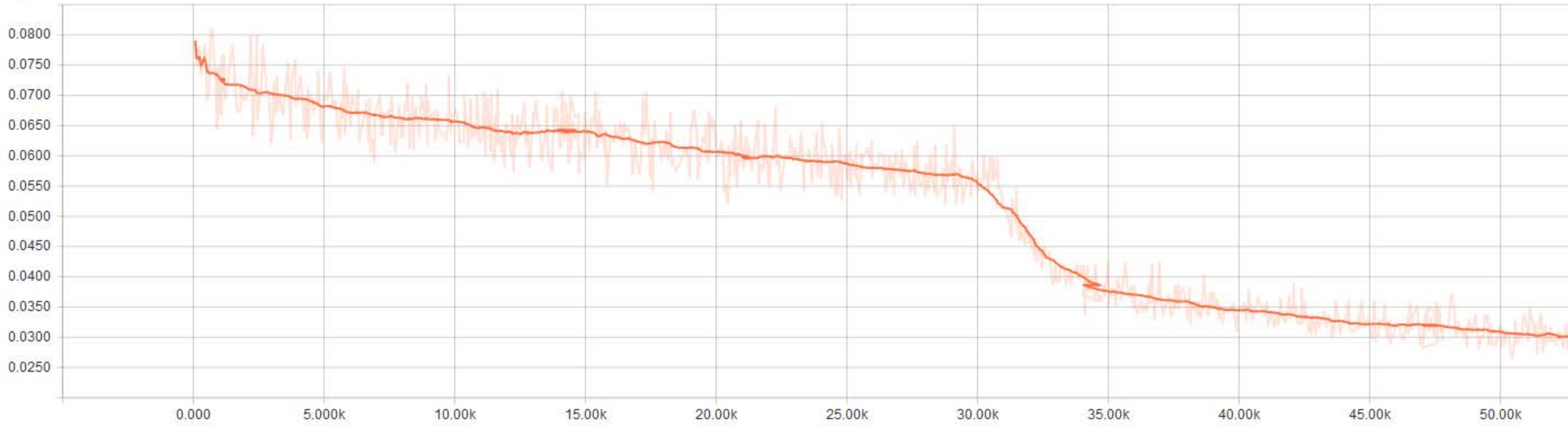}
			\small{\text{(b) Image-to-Video translation on PaSC}}
			\label{fig:side:b}
		\end{minipage}
		
		\begin{minipage}[t]{0.49\linewidth}
			\centering
			\includegraphics[width=1\linewidth]{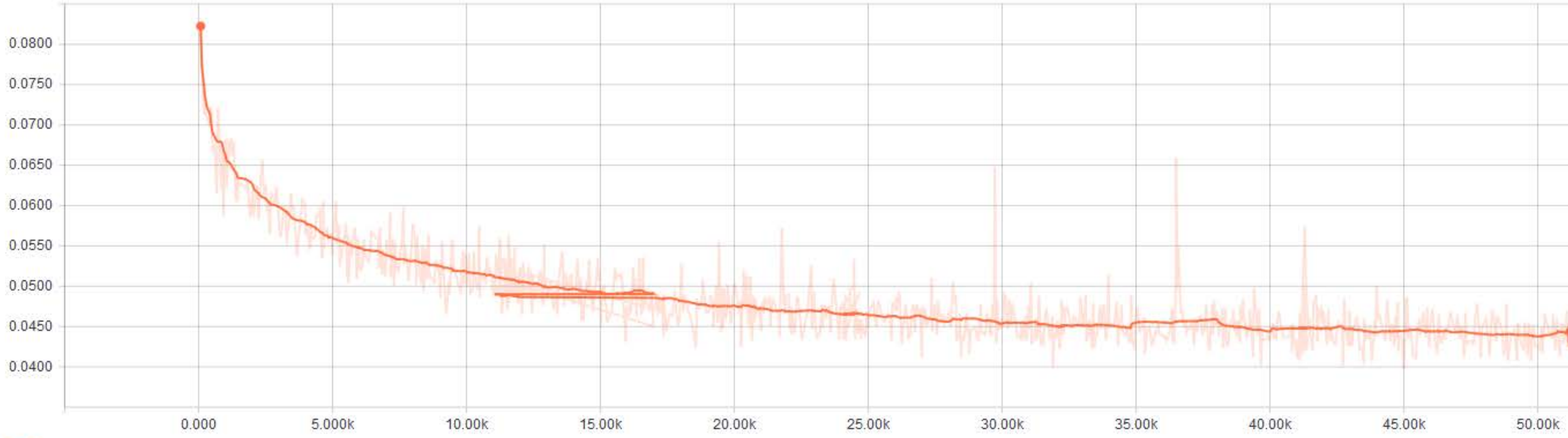}
			\small{\text{(c) Video-to-Image translation on Celeb}}
			\label{fig:side:b}
		\end{minipage}
		\begin{minipage}[t]{0.49\linewidth}
			\centering
			\includegraphics[width=1\linewidth]{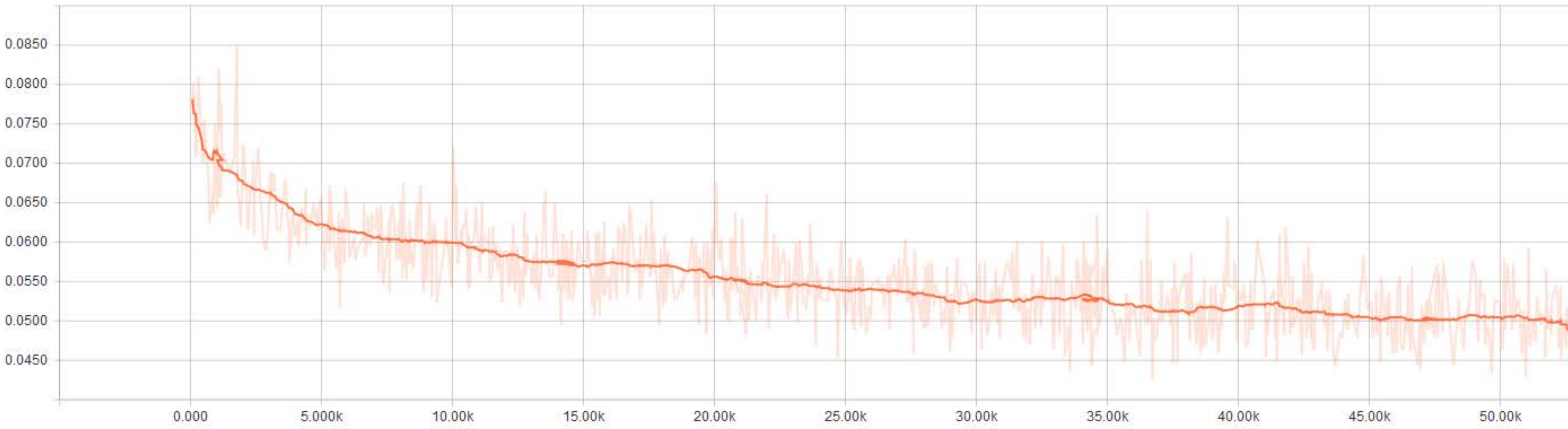}
			\small{\text{(d) Video-to-Image translation on PaSC}}
			\label{fig:side:b}
		\end{minipage}
	\end{center}
	\caption{Identity score curve of the proposed model during network training. Here, the values in x-axis and y-axis respectively denotes the training iteration number and the identity score, which is measured by L2 distance of the FaceNet \cite{schroff2015facenet} face embeddings.}
	\label{fig:idcyclegan_idcurve}
\end{figure*}

\begin{table}[t]
	\small
	\begin{center}
		\begin{tabular}{c|c|c||c|c}
			\hline
			\multirow{2}{*}{FaceNet Score} & \multicolumn{2}{c||}{Image-to-Video} & \multicolumn{2}{c}{Video-to-Image}\\
			\cline{2-5} & Celeb & PaSC & Celeb & PaSC \\
			\hline
			Model1 & 0.0611 & 0.0523 & 0.0675 & 0.0678  \\
			Model2 & 0.0493 & 0.0486 & 0.0621 & 0.0610  \\
			Model3 & 0.0428 & 0.0356 & 0.0538 & 0.0504  \\
			\hline
		\end{tabular}
	\end{center}
	\caption{FaceNet score (measured by L2 distance of the FaceNet \cite{schroff2015facenet} face embeddings) of image-to-video and video-to-image translation for the three proposed models on the Celeb and PaSC datasets.} 
	\label{tab:facenetscore}
\end{table}

\begin{figure*}[t]
	\begin{center}
		\begin{minipage}[t]{0.12\linewidth}
			\centering
			\includegraphics[width=0.8in]{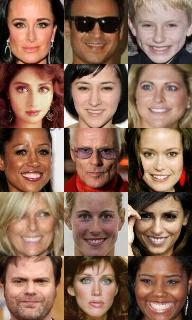}
			\footnotesize{\text{(a) Input Image}}
			\label{fig:side:b}
		\end{minipage}
		\begin{minipage}[t]{0.12\linewidth}
			\centering
			\includemovie[poster=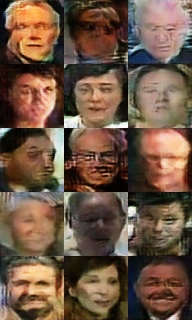, mouse=true, autoplay, repeat=10]{2.05cm}{3.41cm}{cyclegan_celeb_samples_test_avi_vid_50001_gen.avi}
			\footnotesize{\text{(b) Model1 }}
			\label{fig:side:b}
		\end{minipage}
		\begin{minipage}[t]{0.12\linewidth}
			\centering
			\includemovie[poster=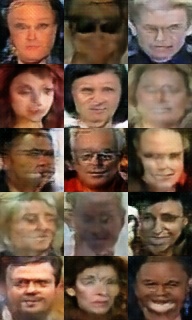, mouse=true, autoplay, repeat=10]{2.05cm}{3.41cm}{cyclegan2_celeb_samples_test_avi_vid_50001_gen.avi}
			\footnotesize{\text{(c) Model2 }}
			\label{fig:side:b}
		\end{minipage}
		\begin{minipage}[t]{0.12\linewidth}
			\centering
			\includemovie[poster=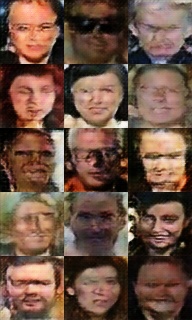, mouse=true, autoplay, repeat=10]{2.05cm}{3.41cm}{idcyclegan_celeb_samples_test_avi_vid_50001_gen.avi}
			\footnotesize{\text{(d) Model3 }}
			\label{fig:side:b}
		\end{minipage}
		\begin{minipage}[t]{0.12\linewidth}
			\centering
			\includemovie[poster=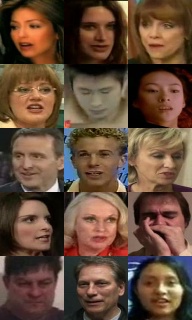, mouse=true, autoplay, repeat=10]{2.05cm}{3.41cm}{idcyclegan_celeb_samples_test_avi_vid_50001_gt.avi}
			\footnotesize{\text{(e) Input Video}}
			\label{fig:side:b}
		\end{minipage}
		\begin{minipage}[t]{0.12\linewidth}
			\centering
			\includegraphics[width=0.8in]{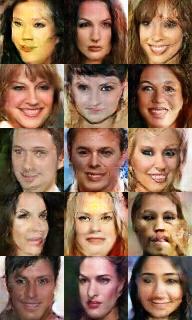}
			\footnotesize{\text{(f) Model1 }}
			\label{fig:side:b}
		\end{minipage}
		\begin{minipage}[t]{0.12\linewidth}
			\centering
			\includegraphics[width=0.8in]{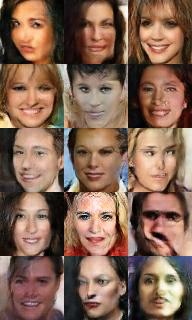}
			\footnotesize{\text{(g) Model2 }}
			\label{fig:side:b}
		\end{minipage}
		\begin{minipage}[t]{0.12\linewidth}
			\centering
			\includegraphics[width=0.8in]{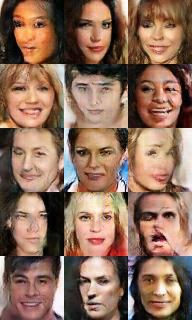}
			\footnotesize{\text{(h) Model3 }}
			\label{fig:side:b}
		\end{minipage}
		
		\begin{minipage}[t]{0.12\linewidth}
			\centering
			\includegraphics[width=0.8in]{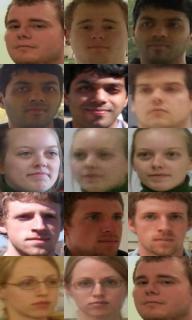}
			\footnotesize{\text{(a) Input Image}}
			\label{fig:side:b}
		\end{minipage}
		\begin{minipage}[t]{0.12\linewidth}
			\centering
			\includemovie[poster=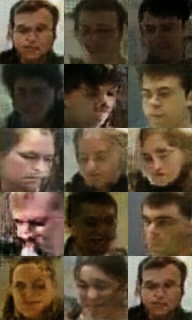, mouse=true, autoplay, repeat=10]{2.05cm}{3.41cm}{cyclegan_pasc_samples_test_avi_vid_50001_gen.avi}
			\footnotesize{\text{(b) Model1 }}
			\label{fig:side:b}
		\end{minipage}
		\begin{minipage}[t]{0.12\linewidth}
			\centering
			\includemovie[poster=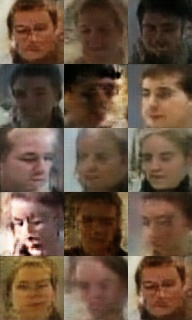, mouse=true, autoplay, repeat=10]{2.05cm}{3.41cm}{cyclegan2_pasc_samples_test_avi_vid_50001_gen.avi}
			\footnotesize{\text{(c) Model2}}
			\label{fig:side:b}
		\end{minipage}
		\begin{minipage}[t]{0.12\linewidth}
			\centering
			\includemovie[poster=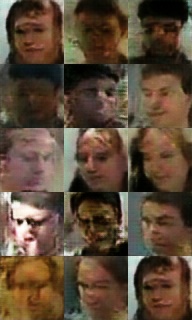, mouse=true, autoplay, repeat=10]{2.05cm}{3.41cm}{idcyclegan_pasc_samples_test_avi_vid_50001_gen.avi}
			\footnotesize{\text{(d) Model3}}
			\label{fig:side:b}
		\end{minipage}
		\begin{minipage}[t]{0.12\linewidth}
			\centering
			\includemovie[poster=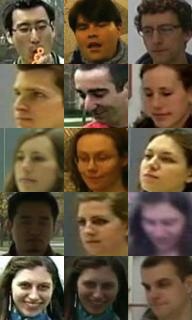, mouse=true, autoplay, repeat=10]{2.05cm}{3.41cm}{idcyclegan_pasc_samples_test_avi_vid_50001_gt.avi}
			\footnotesize{\text{(e) Input Video}}
			\label{fig:side:b}
		\end{minipage}
		\begin{minipage}[t]{0.12\linewidth}
			\centering
			\includegraphics[width=0.8in]{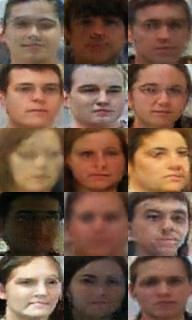}
			\footnotesize{\text{(f) Model1 }}
			\label{fig:side:b}
		\end{minipage}
		\begin{minipage}[t]{0.12\linewidth}
			\centering
			\includegraphics[width=0.8in]{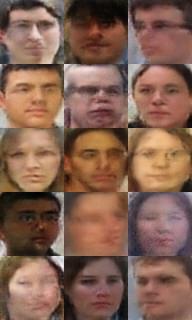}
			\footnotesize{\text{(g) Model2 }}
			\label{fig:side:b}
		\end{minipage}
		\begin{minipage}[t]{0.12\linewidth}
			\centering
			\includegraphics[width=0.8in]{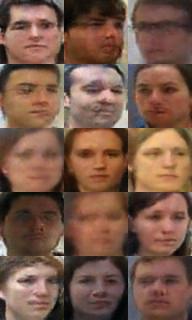}
			\footnotesize{\text{(h) Model3 }}
			\label{fig:side:b}
		\end{minipage}

	\end{center}
	\caption{More image-to-video face translation results (b)-(c)-(d) and video-to-image face translation results (f)-(g)-(h) on the Celeb and PaSC databases. Please notice that the figure is best viewed via the Adobe Acrobat Reader. Click each involved video to play the dynamic video clip.}
	\label{fig:translation_avi}  
\end{figure*}

\subsection{Quantitative Evaluation}

We use verification score of the state-of-the-art face verification network FaceNet \cite{schroff2015facenet} to measure the quantitative results of the generated 1000 samples using the three different translation model. As reported in Table.\ref{tab:facenetscore}, we can find that the proposed Model3 can achieve the lowest verification score which means it has the strongest ability to preserve the identity during the image-to-video and video-to-image face translation task. 

In addition, we also study the identity score curve of the proposed Model3 that explicitly includes the identity preserving cost. As depicted in Fig.\ref{fig:idcyclegan_idcurve}, on the two used datasets, the identity-based cost can always decrease stably as the training iteration number increases towards 50K. This demonstrates that the proposed model can be trained to transfer samples with keeping identity better. 

\section{Conclusion and Future Work}

This paper presented a new problem of unpaired face translation between static images and dynamic videos which can be applied to video face prediction and enhancement. To handle the problem, we proposed a cyclic image-video translation model to bidirectionally translate faces between images and videos in a unified GAN model. To better preserve the identity during translation, we additionally introduce a face verification model that ensures to maintain the facial identity of the original input faces in images or videos. Evaluations on two standard face database demonstrate the proposed model is able to perform image-to-video/video-to-image face translation well in both terms of visual quality and identity maintenance.

The proposed model actually balances the three different terms: GAN loss, cycle consistency loss and identity preserving loss, which actually correspond to the designed critic model, generator model and face verification model. We empirically choose the default setting to make the balance among them. Hence, there may remain an interesting problem to balance such three functions better.

In addition, we actually freeze the network parameters of the pretrained FaceNet during the training process.  Pursuing a better gradient feedback from the face model to better the generated results in both terms of visual appearance and semantic maintenance still remains challenging. It would be also very interesting if we can simultaneously train the generator, discriminator (critic model) and face verification model.

As future work, for better balance between image quality and identity preservation, we will make some efforts to automatically control the hyperparameter between them. Besides, we will also borrow the successful insights of existing stacked GANs \cite{huang2016stacked, han2017stackgan, karras2017progressive} to extend this work to generate  higher-quality images and videos. Lastly, it would be also worth exploring the direction of generalizing the proposed model to translation on video-to-video and even 3D volume data like 3D brain image in medical imaging. 

\section*{Acknowledgements}

We would like to thank NVidia for donating the GPUs used in this work.


\small

\begin{thebibliography}{10}\itemsep=-1pt
	
	\bibitem{karras2017progressive}
	T.~Karras, T.~Aila, S.~Laine, and J.~Lehtinen.
	\newblock Progressive Growing of {GANs} for Improved Quality, Stability, and Variation.
	\newblock {\em arXiv preprint arXiv:1710.10196}, 2017.
	
	\bibitem{arjovsky2017wasserstein}
	M.~Arjovsky, S.~Chintala, and L.~Bottou.
	\newblock Wasserstein generative adversarial networks.
	\newblock In {\em ICML}, 2017.
	
	
	\bibitem{ba2016layer}
	J.~L. Ba, J.~R. Kiros, and G.~E. Hinton.
	\newblock Layer normalization.
	\newblock {\em arXiv preprint arXiv:1607.06450}, 2016.
	
	\bibitem{berthelot2017began}
	D.~Berthelot, T.~Schumm, and L.~Metz.
	\newblock {BEGAN}: Boundary equilibrium generative adversarial networks.
	\newblock {\em arXiv preprint arXiv:1703.10717}, 2017.
	
	\bibitem{beveridge2013challenge}
	J.~R. Beveridge, J.~Phillips, D.~S. Bolme, B.~Draper, G.~H. Givens, Y.~M. Lui,
	M.~N. Teli, H.~Zhang, W.~T. Scruggs, K.~W. Bowyer, P.~J. Flynn, and S.~Cheng.
	\newblock The challenge of face recognition from digital point-and-shoot
	cameras.
	\newblock In {\em BTAS}, 2013.
	
	\bibitem{efros1999texture}
	A.~A. Efros and T.~K. Leung.
	\newblock Texture synthesis by non-parametric sampling.
	\newblock In {\em Computer Vision, 1999. The Proceedings of the Seventh IEEE
		International Conference on}, volume~2, pages 1033--1038. IEEE, 1999.
	
	\bibitem{goodfellow2014generative}
	I.~Goodfellow, J.~Pouget-Abadie, M.~Mirza, B.~Xu, D.~Warde-Farley, S.~Ozair,
	A.~Courville, and Y.~Bengio.
	\newblock Generative adversarial nets.
	\newblock In {\em NIPS}, 2014.
	
	
	\bibitem{gulrajani2017improved}
	I.~Gulrajani, F.~Ahmed, M.~Arjovsky, V.~Dumoulin, and A.~Courville.
	\newblock Improved training of {Wasserstein GANs}.
	\newblock {\em NIPS}, 2017.
	
	\bibitem{hertzmann2001image}
	A.~Hertzmann, C.~E. Jacobs, N.~Oliver, B.~Curless, and D.~H. Salesin.
	\newblock Image analogies.
	\newblock In {\em Proceedings of the 28th annual conference on Computer
		graphics and interactive techniques}, pages 327--340. ACM, 2001.
	
	\bibitem{huang2017beyond}
	R.~Huang, S.~Zhang, T.~Li, and R.~He.
	\newblock Beyond face rotation: Global and local perception {GAN} for
	photo-realistic and identity preserving frontal view synthesis.
	\newblock {\em ICCV}, 2017.
	
	\bibitem{huang2016stacked}
	X.~Huang, Y.~Li, O.~Poursaeed, J.~Hopcroft, and S.~Belongie.
	\newblock Stacked generative adversarial networks.
	\newblock {\em arXiv preprint arXiv:1612.04357}, 2016.
	
	\bibitem{ioffe2015batch}
	S.~Ioffe and C.~Szegedy.
	\newblock Batch normalization: Accelerating deep network training by reducing
	internal covariate shift.
	\newblock In {\em ICML}, 2015.
	
	\bibitem{isola2016image}
	P.~Isola, J.-Y. Zhu, T.~Zhou, and A.~A. Efros.
	\newblock Image-to-image translation with conditional adversarial networks.
	\newblock {\em arXiv preprint arXiv:1611.07004}, 2016.
	
	\bibitem{li2017generative}
	Y.~Li, S.~Liu, J.~Yang, and M.-H. Yang.
	\newblock Generative face completion.
	\newblock {\em CVPR}, 2017.
	
	\bibitem{li2017generate}
	Z.~Li and Y.~Luo.
	\newblock Generate identity-preserving faces by generative adversarial
	networks.
	\newblock {\em arXiv preprint arXiv:1706.03227}, 2017.
	
	\bibitem{liu2017unsupervised}
	M.-Y. Liu, T.~Breuel, and J.~Kautz.
	\newblock Unsupervised image-to-image translation networks.
	\newblock {\em arXiv preprint arXiv:1703.00848}, 2017.
	
	\bibitem{liu2016coupled}
	M.-Y. Liu and O.~Tuzel.
	\newblock Coupled generative adversarial networks.
	\newblock In {\em NIPS}, 2016.
	
	\bibitem{liu2015faceattributes}
	Z.~Liu, P.~Luo, X.~Wang, and X.~Tang.
	\newblock Deep learning face attributes in the wild.
	\newblock In {\em ICCV}, 2015.
	
	\bibitem{lu2017conditional}
	Y.~Lu, Y.-W. Tai, and C.-K. Tang.
	\newblock Conditional {CycleGAN} for attribute guided face image generation.
	\newblock {\em arXiv preprint arXiv:1705.09966}, 2017.
	
	\bibitem{mao2016least}
	X.~Mao, Q.~Li, H.~Xie, R.~Y. Lau, Z.~Wang, and S.~P. Smolley.
	\newblock Least squares generative adversarial networks.
	\newblock {\em ICCV}, 2017.
	
	\bibitem{radford2015unsupervised}
	A.~Radford, L.~Metz, and S.~Chintala.
	\newblock Unsupervised representation learning with deep convolutional
	generative adversarial networks.
	\newblock {\em arXiv preprint arXiv:1511.06434}, 2015.
	
	\bibitem{rosales2003unsupervised}
	R.~Rosales, K.~Achan, and B.~J. Frey.
	\newblock Unsupervised image translation.
	\newblock In {\em ICCV}, 2003.
	
	\bibitem{schroff2015facenet}
	F.~Schroff, D.~Kalenichenko, and J.~Philbin.
	\newblock {FaceNet}: A unified embedding for face recognition and clustering.
	\newblock In {\em CVPR}, 2015.
	
	\bibitem{tran2015learning}
	D.~Tran, L.~Bourdev, R.~Fergus, L.~Torresani, and M.~Paluri.
	\newblock Learning spatiotemporal features with {3D} convolutional networks.
	\newblock In {\em ICCV}, 2015.
	
	\bibitem{tran2017disentangled}
	L.~Tran, X.~Yin, and X.~Liu.
	\newblock Disentangled representation learning {GAN} for pose-invariant face
	recognition.
	\newblock In {\em CVPR}, 2017.
	
	\bibitem{tulyakov2017mocogan}
	S.~Tulyakov, M.-Y. Liu, X.~Yang, and J.~Kautz.
	\newblock {MoCoGAN}: Decomposing motion and content for video generation.
	\newblock {\em arXiv preprint arXiv:1707.04993}, 2017.
	
	\bibitem{vondrick2016generating}
	C.~Vondrick, H.~Pirsiavash, and A.~Torralba.
	\newblock Generating videos with scene dynamics.
	\newblock In {\em NIPS}, 2016.
	
	\bibitem{wolf2011face}
	L.~Wolf, T.~Hassner, and I.~Maoz.
	\newblock Face recognition in unconstrained videos with matched background
	similarity.
	\newblock In {\em CVPR}, 2011.
	
	\bibitem{wu2015light}
	X.~Wu, R.~He, Z.~Sun, and T.~Tan.
	\newblock A light {CNN} for deep face representation with noisy labels.
	\newblock {\em arXiv preprint arXiv:1511.02683}, 2015.
	
	\bibitem{han2017stackgan}
	H.~Zhang, T.~Xu, H.~Li, S.~Zhang, X.~Wang, X.~Huang, and D.~Metaxas.
	\newblock {StackGAN}: Text to photo-realistic image synthesis with stacked
	generative adversarial networks.
	\newblock In {\em {ICCV}}, 2017.
	
	\bibitem{zhang2016joint}
	K.~Zhang, Z.~Zhang, Z.~Li, and Y.~Qiao.
	\newblock Joint face detection and alignment using multitask cascaded
	convolutional networks.
	\newblock {\em IEEE Signal Processing Letters}, 23(10):1499--1503, 2016.
	
	\bibitem{zhao2017multi}
	B.~Zhao, X.~Wu, Z.-Q. Cheng, H.~Liu, and J.~Feng.
	\newblock Multi-view image generation from a single-view.
	\newblock {\em arXiv preprint arXiv:1704.04886}, 2017.
	
	\bibitem{zhao2016energy}
	J.~Zhao, M.~Mathieu, and Y.~LeCun.
	\newblock Energy-based generative adversarial network.
	\newblock {\em arXiv preprint arXiv:1609.03126}, 2016.
	
	\bibitem{zhu2017unpaired}
	J.-Y. Zhu, T.~Park, P.~Isola, and A.~A. Efros.
	\newblock Unpaired image-to-image translation using cycle-consistent
	adversarial networks.
	\newblock {\em ICCV}, 2017.
	
\end{thebibliography}

\end{document}